\documentclass[10pt,onecolumn,draftclsnofoot,journal]{IEEEtran}
\usepackage{amsmath,graphicx,multicol}
\usepackage{amsfonts}
\usepackage{color}
\usepackage{bbm}
\usepackage{cite}
\usepackage{amssymb}
\usepackage{algorithm}
\usepackage{algorithmic}
\usepackage{tabularx}
\usepackage{hyperref}
\usepackage[utf8]{inputenc}
\usepackage[english]{babel}

\usepackage{amsthm}
\usepackage{subcaption}
\usepackage[font={footnotesize}]{caption}
\usepackage{url}
\usepackage{mathtools}
\usepackage{comment}
\usepackage{array,booktabs,dcolumn,calc}
\usepackage{float}
\usepackage[keeplastbox]{flushend}

\newtheorem{definition}{Definition}[]

\DeclareMathOperator*{\argmax}{arg\,max}
\DeclareMathOperator*{\argmin}{arg\,min}

\DeclareMathOperator{\E}{\mathbb{E}}

\def\C{{\mathbf C}}

\def\P{{\mathbf P}}

\def\I{{\mathbf I}}
\def\r{{\mathbf r}}

\def\E{{\mathbf E}}

\def\x{{\mathbf x}}

\def\u{{\mathbf u}}

\def\Z{{\mathbf Z}}
\def\Y{{\mathbf Y}}
\def\r{{\mathbf r}}

\def\A{{\mathbf A}}

\def\P{{\mathbf P}}
\def\U{{\mathbf U}}

\def\R{{\mathbb{R}}}

\def\I{{\mathbf I}}
\def\Z{{\mathbf Z}}

\def\c{{\mathbf c}}

\def\L{{\cal L}}
\def\S{{\cal S}}

\def\X{{\mathbf X}}

\newcommand{\ts}{\textsuperscript}

\let\oldref\ref
\renewcommand{\ref}[1]{(\oldref{#1})}
\newcommand{\RNum}[1]{\uppercase\expandafter{\romannumeral #1\relax}}
\newcolumntype{C}{>{\centering\arraybackslash}b{\widthof{positions}}}
\newcolumntype{d}{D{.}{.}{-2}}
\newcommand{\ra}[1]{\renewcommand{\arraystretch}{#1}}

\makeatletter
\renewcommand{\fnum@figure}{Fig.~\thefigure}
\makeatother

\begin{document}
\title{Evolutionary Self-Expressive Models for \\Subspace Clustering}
\author{Abolfazl~Hashemi,~\IEEEmembership{Student Member,~IEEE,} and Haris~Vikalo,~\IEEEmembership{Senior Member,~IEEE}
\thanks{To appear in IEEE Journal of Selected Topics in Signal Processing, Special Issue on Data Science: Robust Subspace Learning and Tracking, vol. 12, no. 6, December 2018.
	Authors are with the Department of Electrical and Computer Engineering, University of Texas at Austin, Austin, TX 78712 USA (e-mail: \href{mailto:abolfazl@utexas.edu}{abolfazl@utexas.edu}; \href{mailto:hvikalo@ece.utexas.edu}{hvikalo@ece.utexas.edu}).}}
\maketitle
\begin{abstract}
The problem of organizing data that evolves over time into clusters is encountered in a number of practical settings. 
We introduce evolutionary subspace clustering, a method whose objective is to cluster a collection of evolving data
points that lie on a union of low-dimensional evolving subspaces. To learn the parsimonious representation of the 
data points at each time step, we propose a non-convex optimization framework that exploits the self-expressiveness 
property of the evolving data while taking into account representation from the preceding time step. To find an 
approximate solution to the aforementioned non-convex optimization problem, we develop a scheme based on 
alternating minimization that both learns the parsimonious representation as well as adaptively tunes and infers a 
smoothing parameter reflective of the rate of data evolution. The latter addresses a fundamental challenge in 
evolutionary clustering -- determining if and to what extent one should consider previous clustering solutions when 
analyzing an evolving data collection. Our experiments on both synthetic and real-world datasets demonstrate that 
the proposed framework outperforms state-of-the-art static subspace clustering algorithms and existing evolutionary 
clustering schemes in terms of both accuracy and running time, in a range of scenarios.
\end{abstract}
\section{Introduction}\label{sec:intro}
\IEEEPARstart{M}{assive} amounts of high-dimensional data collected by contemporary information 
processing systems create new challenges in the fields of signal processing and machine learning. 
High dimensionality of data presents computational and memory burdens and may adversely affect 
performance of the existing data analysis algorithms. An important unsupervised learning problem
encountered in such settings deals with finding informative parsimonious 
structures characterizing large-scale high-dimensional datasets. This task is critical for  detection 
of meaningful patterns in complex data and enabling accurate and efficient clustering. The problem of 
extracting low-dimensional structures for the purpose of clustering is encountered in many 
applications including motion segmentation and face clustering in computer vision 
\cite{yang2008unsupervised,vidal2008multiframe}, image representation and compression in image 
clustering \cite{ho2003clustering,hong2006multiscale}, robust principal component analysis (PCA), 
and robust subspace recovery and tracking 
\cite{xu2010robust,candes2011robust,yang1995projection,rahmani2017high,rahmani2017randomized}.
In these settings, the data can be thought of as being a collection of points lying on a 
union of low-dimensional subspaces. In addition to having such structural properties, data is often
acquired at multiple points in time. Exploiting the underlying temporal behavior provides more 
informative description and enables improved clustering accuracy. For example, it is well-known that
feature point trajectories associated with motion in a video lie in an affine subspace 
\cite{tron2007benchmark}. Motion during any given short time interval is related to the motion in 
recent past. Therefore, in addition to the union of subspaces structure of the video data, there exists an 
underlying {\it evolutionary structure} characterizing the motion.  Therefore, it is of interest to design and 
investigate frameworks that exploit both 
{\it union of subspaces and temporal smoothness} structures to perform fast and accurate clustering, particularly
in real-time applications where a clustering solution is required at each time step.

In this paper, we formulate and study {\it evolutionary subspace clustering} -- the task of clustering data 
points that lie on a union of {\it evolving} subspaces. We provide a mathematical formulation of 
evolutionary subspace clustering and introduce the {\it convex evolutionary self-expressive model} 
(CESM), an optimization framework that exploits the self-expressiveness property of data and learns 
sparse representations while taking into account prior representations. The task of learning parameters 
of the CESM leads to a non-convex optimization problem which we solve approximately by relying on 
the alternating minimization ideas. In the process of learning data representation, we automatically tune 
a smoothing parameter which characterizes the significance of prior representations, i.e.,
quantifies similarity of the representation in successive time steps. The smoothing parameter is
reflective of the rate of evolution of the data and signifies the amount of temporal changes in
consecutive data snapshots. Note that although we only consider the case of sparse representations, the 
proposed framework can readily be extended to enforce any structures on the learned 
representations, including low rank or low rank plus sparse structures that are often encountered 
in subspace clustering applications. Following extensive simulations on synthetic datasets and 
two real-world datasets originating from real-time motion segmentation (as opposed to offline motion 
segmentation considered in, e.g. \cite{elhamifar2009sparse,elhamifar2013sparse}) and oceanography, 
we demonstrate that the proposed framework significantly improves the performance and 
shortens runtimes of state-of-the-art {\it static} subspace clustering algorithms that only exploit 
the self-expressiveness property of the data.

The rest of this paper is organized as follows. Section \oldref{sec:back} overviews existing approaches 
to subspace clustering and evolutionary clustering. In Section \oldref{sec:pre}, we introduce the evolutionary 
subspace clustering problem and describe the proposed  convex evolutionary self-expressive model. 
Section \oldref{sec:alg} presents algorithms for finding parameters of the CESM. In Section \oldref{sec:ext}, 
we discuss how the proposed framework can be extended to handle issues that often arise 
in practice. Section \oldref{sec:sim} presents experimental results, and the concluding remarks are stated in Section \oldref{sec:concl}. 
\section{Background}\label{sec:back}
In this section, we first define notation used throughout the paper. Then we overview existing subspace 
clustering and evolutionary clustering methods, and highlight distinctive characteristics of evolutionary 
subspace clustering that we introduce and study in the following sections.
\subsection{Notation}
Bold capital letters denote matrices while 
bold lowercase letters represent vectors. Sets as well as subspaces are denoted by calligraphic letters,  
$[n] := \{1,2,\dots,n\}$, and $|\S|$ denotes cardinality of set $\S$. $\X_{ij}$ denotes the $(i,j)$ entry of
$\X$,  $\x_j$ is the $j\ts{th}$ column of $\X$, and $\X_{-j}$ is the matrix constructed by removing the 
$j\ts{th}$ column of $\X$. Additionally, $\X_\S$ is the submatrix of $\X$ that contains the columns of $\X$ 
indexed by the set $\S$. {\color{black}Objects $\X_{t}$, $\x_{t}$, $\mathcal{X}_{t}$, and $x_t$ denote evolving 
matrix, vector, set, and scalar at time $t$, respectively.} $\P_\S^\bot=\I-\X_\S \X_\S^\dagger$ is the projection 
operator onto the orthogonal complement of the subspace spanned by the columns of $\X_\S$, where 
$\X_\S^\dagger=\left(\X_\S^{\top}\X_\S\right)^{-1}\X_\S^{\top}$ denotes the Moore-Penrose pseudo-inverse 
of $\X_\S$ and $\I$ is the identity matrix. Further, $\|\X\|_\ast$ denotes the nuclear norm of $\X$ defined as 
the sum of singular values of $\X$. Finally, $(x)_+$  returns its argument if it is non-negative and returns 
zero otherwise, and $\mathrm{sgn}(x)$ returns the sign of its argument.
\subsection{Subspace clustering}
Subspace clustering has drawn significant attention over the past decade (see, e.g., \cite{vidal2011subspace}
and the references therein). The goal of subspace clustering is to organize data into clusters such that
each cluster collects points that belong to the same subspace. Among various approaches to subspace 
clustering, methods that rely on spectral clustering \cite{ng2001spectral} to analyze the similarity matrix 
which represents relations among data points have received much attention due to their simplicity, 
theoretical rigor, and superior performance. These methods assume that the data is {\em self-expressive} 
\cite{elhamifar2009sparse}, i.e., each data point can be represented by a linear combination of other 
points in the union of subspaces. The self-expressiveness property of data motivates the search for a so-called 
subspace preserving similarity matrix that is reflective of similarities among data points originating from the 
same subspace. To form the similarity matrix, the sparse subspace clustering (SSC) method in 
\cite{elhamifar2009sparse,elhamifar2013sparse} employs a sparse reconstruction algorithm known as basis 
pursuit (BP) which aims to minimize an $\ell_1$-norm objective by means of convex optimization techniques 
such as the interior point method \cite{kim2007interior} or alternating direction method of multipliers (ADMM) 
\cite{boyd2011distributed}. In \cite{dyer2013greedy,you2015sparse}, orthogonal matching pursuit (OMP) is 
used to greedily build the similarity matrix while in 
\cite{tschannen2016noisy,chen2018active,park2014greedy,hashemi2017accelerated,rahmani2017innovation} 
the similarity matrix is constructed by exploring different selection criteria. Low rank representation (LRR) 
subspace clustering approaches in \cite{lu2012robust,liu2013robust,favaro2011closed,vidal2014low} perform 
convex optimization of objective functions that consist of $\ell_2$-norm and nuclear norm regularization terms
and build the similarity matrix via singular value decomposition (SVD) of data. {\color{black} Feng et al. 
\cite{feng2014robust} search for a block-diagonal 
similarity matrix capturing relations among data points that lie on a union of subspaces. In the scenarios where 
self-expressive data can be represented by multiple distinct feature sets, multi-view subspace clustering 
\cite{gao2015multi} attempts to perform subspace clustering on each view simultaneously, while providing 
guarantees of the consistence of clustering structures associated with different views. In 
\cite{nie2016subspace}, the task of low-rank representation learning and segmentation of data is performed 
jointly by identifying individually low-rank segmentations and exploiting the Schatten $p$-norm relaxation of the 
non-convex rank objective function.}
Finally, \cite{heckel2015robust} presents an algorithm that constructs the similarity matrix via thresholding the 
correlations among data points. 

Let $\X_{t}$ and $\C_t$ denote the data and representation matrices at time $t$, respectively. At their core, all
self-expressive subspace clustering schemes attempt to solve variants of the optimization problem
\begin{equation}\label{eq:static}
\begin{aligned}
& \underset{\C_t}{\text{min}}
\; \|\C_t\|
& \text{s.t.}\hspace{0.5cm}  \|\X_{t}- \X_{t} \C_t\|_F^2\leq \epsilon, \quad \mathrm{diag}(\C_t) = \mathbf{0},
\end{aligned}
\end{equation}
where, for instance, the norm in the objective function is $\|\cdot\|_1$, 
$\|\cdot\|_0$, and $\|\cdot\|_\ast$ for SSC-BP, SSC-OMP, and LRR schemes, 
respectively, and $\epsilon$ is a predefined threshold that determines to what extend a representation matrix 
$\C_t$ should preserve self-expressiveness of $\X_{t}$. One then defines an affinity (or similarity) matrix 
$\A_t = |\C_t|+|\C_t|^\top$ and applies spectral clustering \cite{ng2001spectral} to 
find the clustering solution.

Performance of self-expressiveness-based subspace clustering schemes was analyzed in various settings. 
It was shown in \cite{elhamifar2009sparse,elhamifar2013sparse} that when the subspaces are disjoint 
(independent), the BP-based method is subspace preserving. Authors of
\cite{soltanolkotabi2012geometric,soltanolkotabi2014robust} take a geometric point of view to further study 
the performance of BP-based SSC algorithm in the setting of intersecting subspaces and in the presence 
of outliers. These results are extended to the OMP-based SSC \cite{dyer2013greedy,you2015sparse} and 
matching pursuit-based SSC \cite{tschannen2016noisy}. 

Recently, further extensions of SSC and LRR frameworks were developed. In particular, an SSC-based 
approach that jointly performs representation learning and clustering is proposed in \cite{li2017structured} 
while the authors of 
\cite{elhamifar2016high,li2016structured,fan2017sparse,charles2017subspace,tsakiris2018theoretical} 
extend the SSC framework to handle datasets with missing information. Time complexity and memory 
footprint challenges of the LRR framework motivated the development of its online counterpart in 
\cite{shen2016online}. The temporal subspace clustering scheme \cite{li2015temporal} assumes that one 
data point is sampled at each time step and sets the goal of grouping the data points into sequential 
segments, followed by clustering the segments into their respective subspaces. However, neither of 
these approaches considers the possibility of an evolutionary structure in the {\it feature space}, the 
setting studied in the current paper. Instead, prior works assume that the data points are received in an 
online fashion (as opposed to having evolving features) and, once acquired, are fixed and do not evolve 
with time. Therefore, just as the original SSC and LRR frameworks, the subsequent variants of subspace
clustering can be categorized as being {\it static}. In contrast, the evolutionary subspace clustering 
problem studied in the current paper is focused on improving clustering quality by judiciously combining 
parsimonious representations from multiple time steps while exploiting the union of subspaces structure 
of the data.

A related problem to subspace clustering is that of robust principal component analysis (PCA) and robust 
subspace recovery and tracking 
\cite{xu2010robust,candes2011robust,yang1995projection,rahmani2017high,rahmani2017randomized}. 
There, the goal is to identify outliers (which in some applications may actually be the objects of interest) to 
perform PCA and find a {\it single} low-dimensional subspace which best fits a collection of points taken 
from a high-dimensional space. State-of-the-art methods perform this task by decomposing the data 
matrix into a sum of low rank and sparse matrices. Note that, in robust subspace recovery, the data matrix 
consists of all the snapshots of data which are assumed to lie on a single subspace (except for outliers). 
Therefore, this problem, too, is inherently different from the evolutionary subspace clustering framework 
that we study in the current paper.
\subsection{Evolutionary clustering} \label{sec:back-evo}
The topic of evolutionary clustering has attracted significant attention in recent years 
\cite{chakrabarti2006evolutionary,xu2014adaptive,folino2014evolutionary,arzeno2017evolutionary}. The 
problem was originally introduced in \cite{chakrabarti2006evolutionary} where the authors proposed a 
framework for evolutionary clustering by adding a temporal smoothness penalty to a static clustering 
objective. Evolutionary extensions of agglomerative hierarchical clustering and k-means were presented 
as examples of the general framework. Evolutionary clustering has been applied in a variety of practical 
settings such as tracking in dynamic networks \cite{czink2007tracking,folino2014evolutionary} and study 
of climate change \cite{gunnemann2012tracing}, generally improving the performance of static clustering 
algorithms. Non-parametric Bayesian evolutionary clustering schemes employing hierarchical Dirichlet 
process are developed in \cite{ahmed2008dynamic,xu2008evolutionary,ahmed2012timeline}. An 
evolutionary affinity propagation clustering algorithm that relies on message passing between the nodes 
of an appropriately defined factor graph is developed in \cite{arzeno2017evolutionary}. Chi et al. 
\cite{chi2007evolutionary,chi2009evolutionary} proposed two frameworks for evolutionary spectral clustering 
referred to as preserving cluster quality (PCQ) and preserving cluster membership (PCM) schemes. In the 
PCQ formulation, the temporal cost at time $t$ is determined based on the quality of the partition formed 
using data from time $t-1$; in PCM, the temporal cost is a result of comparing the partition at time $t$ with 
the partition at $t-1$. The authors of \cite{rosswog2008detecting} proposed evolutionary extensions of 
k-means and agglomerative hierarchical clustering by filtering the feature vectors using a finite impulse 
response filter which combines the measurements of feature vectors and uses them to find an affinity matrix 
for clustering. Their approach essentially tracks clusters across time by extending the similarity between 
points and cluster centers to include their positions at previous time steps. However, the method in 
\cite{rosswog2008detecting} is limited to the settings where the number of clusters does not change with 
time. Following the idea of modifying similarities followed by static clustering, Xu et al. proposed AFFECT, 
an evolutionary clustering method where the matrix indicating similarity between data points at a given time 
step is assumed to be the sum of a deterministic matrix (the affinity matrix) and a Gaussian noise matrix 
\cite{xu2014adaptive}.

To  find clustering solutions at multiple points in time for evolutionary data characterized by union-of-subspaces 
structure, one might consider concatenating the data snapshots from the first until the current time 
instance and performing subspace clustering on such a set. In this approach, which we refer to as 
concatenate-and-cluster (C\&C), finding the clustering solution at time $t$ would involve forming the matrix 
$\bar{\X}_t = {[{\X}_1^\top,\dots,{\X}_t^\top]}^\top$ and performing clustering $\bar{\X}_t$ via subspace 
clustering approaches. However, due to a significant increase in the number of features (caused by data 
concatenation), such a procedure would incur computational complexity that grows with time (depending on 
the subspace clustering method, the complexity would be either quadratic or cubic in time). Perhaps more 
importantly, the C\&C approach lacks the ability to discover subtle temporal changes in data organization and 
attempts to fit a clustering solution to a single union-of-subspaces structure; in other words, clustering of 
concatenated data fails to account for temporal evolution of subspaces.

As an illustrative example, consider the task of {\it real-time} motion segmentation \cite{smith1995asset,collins2005online} 
where the goal is to identify and track motions in a video sequence. Real-time motion segmentation is related to the 
{\it offline} motion segmentation task studied in \cite{elhamifar2009sparse,elhamifar2013sparse}. The difference 
between the two is that in the offline setting clustering is performed once, after receiving all the frames in the sequence, 
while in the real-time setting clustering steps are performed after receiving each snapshot of data (See Section 
\ref{sec:sim-motion}). 
The subspaces representing the motions evolve; while subspaces in subsequent snapshots are similar, those that are
associated with snapshots separated more widely in time may be drastically different. For this reason, imposing a 
single structure, as in the aforementioned C\&C approach, may lead to poor clustering solutions. Therefore, a scheme 
that judiciously exploit the evolutionary structure while acknowledging the union-of-subspaces structure is needed.

Let $\A_{t-1}$ and $\bar{\A}_{t}$ denote the affinity matrix at time $t-1$ and the affinity matrix constructed 
solely from $\X_{t}$, respectively. State-of-the-art evolutionary clustering algorithms, e.g., 
\cite{chi2007evolutionary,chi2009evolutionary,rosswog2008detecting,xu2014adaptive}, apply a static 
clustering algorithm such as spectral clustering to process the following affinity matrix
\begin{equation}\label{eq:pre}
	\A_{t} = \alpha_t\bar{\A}_{t} + (1-\alpha_t)\A_{t-1},
\end{equation}
where $\alpha_t$ is the so-called smoothing parameter at time $t$. The affinity matrix $\bar{\A}_{t}$ is 
typically constructed from $\X_{t}$ using general similarity notions such as the negative Euclidean 
distance of the data points or its exponential variant. 

The recursive construction of the affinity matrix shown above brings up several questions. First, note that 
when $\bar{\A}_{t}$ is determined from \eqref{eq:pre}, one does not take into account representation of data 
points in previous time steps; as we show in our experimental studies, this may lead to poor performance in 
subspace clustering applications. More importantly, apart from the AFFECT algorithm \cite{xu2014adaptive}, 
none of the existing evolutionary clustering schemes provides a procedure for finding the smoothing 
parameter $\alpha_t$ which determines how much weight is placed on historic data. Instead, existing methods 
typically set $\alpha_t$ according to the user's preference for the temporal smoothness of the clustering 
results. AFFECT relies on an iterative shrinkage estimation approach to automatically tune $\alpha_t$. However, 
to find the smoothing parameter, AFFECT makes certain strong assumptions on the structure of the affinity matrix. 
In particular, it assumes a block structure that holds only if the data at each time $t$ is a realization of a dynamic 
Gaussian mixture model, which is typically not the case in practice, especially in subspace clustering applications 
such as motion segmentation. Indeed, as our simulation results demonstrate, typical values of smoothing 
parameter found by the shrinkage estimation approach of AFFECT in motion segmentation application is 
$\alpha_t \approx 0.5$ regardless of whether the data is static or evolutionary. This is counterintuitive since, 
e.g., for static data we expect $\alpha_t \approx 0$.

To address the above challenges, we develop a novel framework for clustering temporal high-dimensional data 
that contains points lying on a union of low-dimensional subspaces. The proposed framework exploits the 
self-expressiveness property of data to learn a representation for $\X_{t}$ while at the same time takes into 
account data representation learned in the previous time step. Moreover, we propose a novel strategy that relies 
on alternating minimization to automatically learn the smoothing parameter $\alpha_t$ at each time step. As our 
extensive simulation results demonstrate, the smoothing parameter inferred by the proposed CESM framework 
captures temporal behavior and adapts to sudden changes in data. Therefore, the smoothing parameter found
by the proposed framework is reflective of the rate of data evolution and quantifies the significance of prior 
representations when clustering data at time $t$. Note that even though in this paper we focus on evolutionary 
self-expressive models with sparse representation, the proposed framework can be extended in straightforward 
manner to include other representation learning frameworks such as LRR.
\section{Evolutionary Subspace Clustering}\label{sec:pre}
Let $\{\x_{t,i}\}_{i=1}^{N_t}$ be a collection of (evolving) real-valued $D_t$-dimensional data points at 
time $t$ and let us organize those points in a matrix $\X_t = [\x_{t,1},\dots,\x_{t,N_t}] \in \R^{D_t\times N_t}$. 
The data points are drawn from a union of $n_t$ evolving subspaces $\{\S_{t,i}\}_{i=1}^{n_t}$ with 
dimensions $\{d_{t,i}\}_{i=1}^{n_t}$. Without a loss of generality, we assume that the columns of $\X_t$, i.e., 
the data points, are normalized vectors with unit $\ell_2$ norm. \footnote{As we proceed, for the simplicity 
 of notation we may omit the time index.} Due to the underlying  union of subspaces structure, 
 the data points satisfy the self-expressiveness property \cite{elhamifar2009sparse} 
 formally stated below.
\begin{definition}
	\textit{A collection of evolving data points $\{\x_{t,i}\}_{i=1}^{N_t}$ satisfies the self-expressiveness 
	property if each data point has a linear representation in terms of the other points in the collection, i.e., 
	there exist a representation matrix $\C_t$ such that}
	\begin{equation}\label{def:1}
	\X_t = \X_t \C_t, \quad \mathrm{diag}(\C_t) = \mathbf{0}.
	\end{equation}
\end{definition}

The goal of subspace clustering is to partition $\{\x_{t,i}\}_{i=1}^{N_t}$ into $n_t$ groups such that the 
data points that belong to the same subspace are assigned to the same cluster. To distinguish between 
different methods, we refer to subspace clustering schemes that find a representation matrix $\C_t$ 
which satisfies \eqref{def:1} as the {\it static subspace clustering} methods. As stated in Sections 
\ref{sec:intro} and \ref{sec:back}, in many applications the subspaces and the data points lying on the 
union of those subspaces evolve over time. Imposing the self-expressiveness property helps exploit 
the fact that the data points belong to a union of subspaces. However, \eqref{def:1} alone does not 
capture potential evolutionary structure of the data. To this end, we propose to find a representation 
matrix $\C_t$, for each time $t$, such that
\begin{equation}\label{eq:mg}
\C_t = f_\theta(\C_{t-1}), \quad \X_t = \X_t \C_t, \quad \mathrm{diag}(\C_t) = \mathbf{0}.
\end{equation} 
In other words, the representation matrix $\C_t$ is assumed to be {\color{black} a matrix-valued function 
parametrized by $\theta$} that captures the self-expressiveness property of data while also promoting a
relation to the representation matrix at a preceding time instance, $\C_{t-1}$. The function 
$f_\theta:\mathcal{P}_\C \rightarrow \mathcal{P}_\C $ may in principle be any appropriate parametric 
function while the set $\mathcal{P}_\C \subseteq \R^{N\times N}$ stands for any preferred parsimonious 
structures imposed on the representation matrices at each time instant, e.g., sparse or low-rank 
representations. We refer to subspace clustering schemes that satisfy \eqref{eq:mg} as {\it evolutionary 
subspace clustering} methods. To find such a representation matrix $\C_t$, we formulate and solve the 
optimization
\begin{equation}\label{eq:mg2}
\begin{aligned}
& \underset{\theta}{\text{min}}
\quad \|\X_t - \X_t f_\theta(\C_{t-1})\|_F^2 \\
& \text{s.t.}\hspace{0.5cm}  {\color{black}f_{\theta}(\C_{t-1}) \in \mathcal{P}_\C},\\
\end{aligned}
\end{equation}
{\color{black}{and use the resulting representation matrix $\C_t = f_{\theta}(\C_{t-1})$ to segment the data.}}

The evolutionary subspace clustering problem \eqref{eq:mg2} is essentially a general constrained 
representation learning problem. Given any combination of $(f_\theta,\mathcal{P}_\C)$, a solution to 
\eqref{eq:mg2} results in a distinct evolutionary subspace clustering framework. After finding a 
solution to \eqref{eq:mg2} and setting $\C_t = f_{\theta^\ast}(\C_{t-1})$, we construct an affinity 
matrix $\A_t = |\C_t|+|\C_t|^\top$ and then apply spectral clustering to $\A_t$.

In this paper, we restrict our studies to the case where $\mathcal{P}_\C$ is the set of sparse representation 
matrices and consider a simple and interpretable form of the parametric function $f_\theta$. Other structures 
and more complex parametric functions are left for future work.
\subsection{Convex evolutionary self-expressive model}
Consider the function
\begin{equation}\label{eq:cesmf}
	\C_{t} = f_{\theta}(\C_{t-1}) = \alpha\U+(1-\alpha)\C_{t-1},
\end{equation}
where the values of parameters $\theta = (\U,\alpha)$ specify the relationship between $\C_{t-1}$ and 
$\C_{t}$, and need to be learned from data. Intuitively, the {\it innovation representation matrix} $\U$ 
captures changes in the representation of data points between consecutive time steps. The other term on 
the right-hand side of \eqref{eq:cesmf}, $(1-\alpha)\C_{t-1}$, is the part of temporal representation that 
carries over from the previous snapshot of data. Therefore, the parametric function in  \eqref{eq:cesmf} 
assumes that the representation at time $t$ is a convex combination of the representation at $t-1$, 
$\C_{t-1}$, and the ``innovation'' in the representation captured by matrix $\U$. Parameter 
$0\leq\alpha\leq 1$ quantifies significance of the previous representation on the structure of data points 
at time $t$ (i.e., it is reflective of the ``memory" of representations). Intuitively, if the data is static we 
expect parameters to take on the values ($\alpha = 0$, $\U = \mathbf{0}$) or ($\alpha = 1$, $\U = \C_{t-1}$). 
Conversely, if the temporally evolving data is characterized by a subspace structure that undergoes 
significant changes, we expect $\alpha$ to be relatively close to $0.5$.

Since at each time step we seek a sparse self-representation of data, the innovation matrix $\U$ should
be sparse and satisfy $\mathrm{diag}(\U) = \mathbf{0}$. Therefore, for the evolutionary model
\eqref{eq:cesmf}, search for the best collection of parameters that relate $\C_{t-1}$ and $\C_{t}$ leads to 
optimization
\begin{equation}\label{eq:cesm}
\begin{aligned}
& \underset{\U,\alpha}{\text{min}}
\quad \|\X_t - \X_t (\alpha\U+(1-\alpha)\C_{t-1})\|_F^2 \\
& \text{s.t.}\hspace{0.5cm}  \mathrm{diag}(\U) = \mathbf{0}, \quad \|\U\|_0\leq k,\\
&\hspace{0.9cm} 0\leq\alpha\leq 1.
\end{aligned}
\end{equation}
In the above optimization, $k$ determines sparsity level of the innovation. Since each point in $\S_i$ can 
be expressed in terms of at most $d$ points in $\S_i$, we typically set $k\leq d$.

We refer to \eqref{eq:cesm} as the convex evolutionary self-expressive model (CESM) for the evolutionary 
subspace clustering. Note that due to the cardinality constraint, \eqref{eq:cesm} is a non-convex optimization 
problem. In Section \ref{sec:alg}, we present methods that rely on alternating minimization to efficiently find 
an approximate solution to  (\ref{eq:cesm}).

\section{Alternating Minimization Algorithms for Evolutionary Subspace Clustering}\label{sec:alg}
In this section, we present alternating minimization schemes for finding the innovation representation 
matrix $\U$ and smoothing parameter $\alpha$, i.e., for solving (\ref{eq:cesm}).
\subsection{Finding parameters of the CESM model}
We solve \eqref{eq:cesm} for $\U$ and $\alpha$ in an alternating fashion. In particular, given $\U_{t-1}$, 
the innovation representation matrix found at time $t-1$, we determine value of the smoothing parameter 
according to 
\begin{equation}\label{eq:cesma}
\alpha = \argmin_{0\leq\bar{\alpha}\leq 1} \|\X_t - \X_t (\bar{\alpha}\U_{t-1}+(1-\bar{\alpha})\C_{t-1})\|_F^2.
\end{equation} 
The objective function in \eqref{eq:cesma} is unimodal and convex; in our implementation, we rely on 
the golden-section search algorithm \cite{kiefer1953sequential} to efficiently find $\alpha$. Having found 
$\alpha$, we arrive at the representation learning step which requires solving
\begin{equation}\label{eq:cesmu}
\begin{aligned}
& \underset{\U}{\text{min}}
\quad \|\X_t - \X_t (\alpha\U+(1-\alpha)\C_{t-1})\|_F^2 \\
& \text{s.t.}\hspace{0.5cm}  \mathrm{diag}(\U) = \mathbf{0}, \quad \|\U\|_0\leq k,
\end{aligned}
\end{equation}
which is a non-convex optimization problem due to the cardinality constraint. Let 
$\widetilde{\X}_t = \frac{1}{\alpha}(\X_t-(1-\alpha)\X_t\C_{t-1})$. Then, \eqref{eq:cesmu} can equivalently be written as
\begin{equation}\label{eq:cesmu2}
\begin{aligned}
& \underset{\U}{\text{min}}
\quad \|\widetilde{\X}_t - \X_t \U\|_F^2 \\
& \text{s.t.}\hspace{0.5cm}  \mathrm{diag}(\U) = \mathbf{0}, \quad \|\U\|_0\leq k.
\end{aligned}
\end{equation}
The optimization problem \eqref{eq:cesmu2} is clearly related to static subspace clustering with sparse 
representation (cf. \eqref{eq:static}) and, in general, to compressed sensing problems 
\cite{donoho2006compressed}. Similar to static sparse subspace clustering schemes 
\cite{elhamifar2009sparse,elhamifar2013sparse,dyer2013greedy,you2015sparse,hashemi2017accelerated}, 
one can employ compressed sensing approaches such as basis pursuit (BP) \cite{chen2001atomic} 
(or the related LASSO \cite{tibshirani1996regression}), orthogonal matching pursuit (OMP) 
\cite{pati1993orthogonal}, and orthogonal least squares (OLS) \cite{chen1989orthogonal} algorithms to 
find a suboptimal innovation matrix $\U$ in polynomial time. 

In particular, the basis pursuit representation learning strategy leads to the convex program  
\begin{equation}\label{eq:cesmubp}
\begin{aligned}
& \underset{\U}{\text{min}}
\quad \|\U\|_1+\frac{\lambda}{2}\|\widetilde{\X}_t - \X_t \U\|_F^2 \\
& \text{s.t.}\hspace{0.5cm}  \mathrm{diag}(\U) = \mathbf{0},
\end{aligned}
\end{equation}
which can be solved using any conventional convex solver (see Section \ref{sec:ext} for an 
ADMM-based implementation). Here, $\lambda>0$ is a regularization parameter that determines 
sparsity level of the innovation representations.

For the OMP-based strategy, to learn the representation for each data point $\x_{t,j}$, $j \in [N]$, we 
define an initial residual vector $\r_0 = \widetilde{\x}_{t,j}$ and greedily select $k$ data points indexed 
by $\mathcal{A}_k = \{i_1,\dots,i_k\} \subset [N]$ that contribute to the innovation representation of 
$\x_{t,j}$ according to 
\begin{equation}\label{eq:aesmuomp}
i_\ell = \argmax_{i\in[N]\backslash\mathcal{A}_{\ell-1}\cup\{j\} } |\r_{\ell-1}\x_{t,i}|^2, 
\end{equation}
where $ \ell \in [k]$. The residual vector is updated according to 
$\r_\ell =  \P^\bot_{\mathcal{A}_\ell}\widetilde{\x}_{t,j}$, where $\P_{\mathcal{A}_\ell}$ is the projection 
operator onto the subspace spanned by $\X_{t,{\mathcal{A}_\ell}}$ (i.e., the columns of $\X_t$ that are 
indexed by $\mathcal{A}_\ell$). 
Once $\mathcal{A}_k$ is determined, the innovation representation is computed as the 
least square solution $\u_j = \X_{t,{\mathcal{A}_k}}^\dagger \widetilde{\x}_{t,j}$.

The OLS-based representation learning strategy is similar to that of OMP, except the selection criterion 
is modified to
\begin{equation}\label{eq:aesmuols}
i_\ell = \argmax_{i\in[N]\backslash\mathcal{A}_{\ell-1}\cup\{j\} } \frac{|\r_{\ell-1}\x_{t,i}|^2}{\|\P^\bot_{\mathcal{A}_{\ell-1}}\x_{t,i}\|_2^2}. 
\end{equation}
Finally, \eqref{eq:cesmf} yields the desired representation matrix $\C_t$. 
\subsection{Complexity analysis}
The computational complexity of the proposed alternating minimization schemes is analyzed next. 

Since it takes $\mathcal{O}(N^2)$ to evaluate the objective functions in \eqref{eq:cesma}, the complexity of 
finding the smoothing parameter using the golden-section search or any other linearly convergent optimization 
algorithm is $\mathcal{O}(N^2)$. 

The computational cost of using BP-based strategy to learn the innovation representation matrix $\U$ in 
$\tau$ iterations of the interior-point method is $\mathcal{O}(\tau D N^3)$. However, as we demonstrate in 
Section \ref{sec:ext}, by using an efficient ADMM implementation the complexity can be reduced to 
$\mathcal{O}(\tau_m D^2N^2)$ where $\tau_m$ denotes the maximum number of iterations of the ADMM 
algorithm. 

Since they require search over $\mathcal{O}(N)$ $D$-dimensional data points in $k$ iterations, the
complexity of learning innovation representation matrix using OMP and OLS methods is 
$\mathcal{O}(kDN^2)$ and $\mathcal{O}(kD^2N^2)$, respectively. In Section \ref{sec:ext} we discuss how 
one can reduce the complexity of OMP and OLS-based representation learning methods to 
$\mathcal{O}(DN^2)$ using accelerated and randomized greedy strategies.

\section{Practical Extensions}\label{sec:ext}
Here we discuss potential practical issues and challenges that may come up in applications, 
and demonstrate how the proposed frameworks can be extended to handle such cases.
\subsection{Tracking the evolution of clusters}
The CESM framework promotes consistent assignment of data points to clusters over time. However, 
subspaces and the corresponding clusters evolve and thus one still faces the challenge of matching the 
clusters formed at consecutive time steps. This task essentially entails searching over permutations of 
clusters at time $t$ and identifying the one that best matches the collection of clusters at time $t-1$. 
Quality of a matching (i.e., the weight of a matching) is naturally quantified by the number of data points 
common to the pairs of matched clusters. The solution to the so-called maximum weight matching 
problem can be found in polynomial time using the well-known Hungarian algorithm 
\cite{kuhn1955hungarian}, or its variants that handle more sophisticated cases such as one-to-many and 
many-to-one maximum weight matching \cite{greene2010tracking,brodka2013ged}. In our numerical studies, 
we use the Hungarian algorithm to match clusters across time and evaluate clustering accuracy in 
experiments where the ground truth is known.
\subsection{Adding and removing data points over time}
In practice, it may happen that some of the data points vanish over time while new data points are 
introduced. In such settings, the number of data points and hence the dimension of representation 
matrices varies over time. Our proposed framework readily deals with such scenarios, as explained 
next.

Let $\mathcal{T}$ denote the set of indices of data points introduced at time $t$ that were 
not present at time $t-1$. To incorporate these points into the model, we expand $\C_{t-1}$ 
by inserting all-zero vectors in rows and columns indexed by $\mathcal{T}$. New data points 
do not play a role in the temporal representations of other data points but they may participate 
in the innovation representation matrix (i.e., $\U$). Finally,
let $\overline{\mathcal{T}}$ denote the set of indices of data points that were present at time 
$t-1$ but have vanished at time $t$; those points are removed from the model by excluding 
rows and columns of $\C_{t-1}$ indexed by $\overline{\mathcal{T}}$.
\subsection{Accelerated representation learning}
The most computationally challenging step of the proposed evolutionary self-expressive model 
is the representation learning step, i.e., the task of computing the innovation representation matrix 
$\U$. Therefore, when handling evolutionary data containing a large number of high-dimensional 
data points, efficient representation learning methods are needed. To this end, we here discuss 
how to employ BP, OMP, and OLS-based strategies in an accelerated fashion.  

\subsubsection{BP-based representation learning}
We first develop an ADMM algorithm for finding the innovation matrix $\U$ in \eqref{eq:cesmubp} 
following a similar approach to that of \cite{elhamifar2013sparse}.

Define $\widetilde{\X}_t=\frac{1}{\alpha}(\X_t-(1-\alpha)\X_t\C_{t-1})$. 
Introduce an auxiliary matrix $\Z$ and consider the optimization
\begin{equation}\label{eq:admm1}
\begin{aligned}
& \underset{\U,\Z}{\text{min}}
\quad \|\U\|_1+\frac{\lambda}{2}\|\widetilde{\X}_t - \X_t \Z\|_F^2 \\
& \text{s.t.}\hspace{0.5cm}  \Z=\U-\mathrm{diag}(\U),
\end{aligned}
\end{equation}
which is equivalent to the optimization problem \eqref{eq:cesmubp} considered in Section 
\ref{sec:alg}. Form the augmented Lagrangian of \eqref{eq:admm1} to obtain
\begin{equation}\label{eq:admm2}
\begin{aligned}
& \L_\rho(\U,\Z,\Y)=\|\U\|_1+\frac{\lambda}{2}\|\widetilde{\X}_t - \X_t \Z\|_F^2\\
&\qquad\qquad+\frac{\rho}{2}\|\Z-\U+\mathrm{diag}(\U)\|_F^2\\
&\qquad\qquad+\mathrm{tr}(\Y^\top(\Z-\U+\mathrm{diag}(\U))),
\end{aligned}
\end{equation}
where $\rho>0$ and $\Y$ are the so-called penalty parameter and dual variable, respectively. 
Since adding the penalty term makes the objective function \eqref{eq:admm2} strictly convex 
in the optimization variables, we can apply ADMM to solve it efficiently. The ADMM consists of 
the following iterations:
\begin{itemize}
\item $\Z^{\ell+1} = \min_{\Z^{\ell}}\L_\rho(\U^{\ell},\Z^{\ell},\Y^{\ell})$.

According to \cite{boyd2011distributed,elhamifar2013sparse}, this problem has a closed-form 
solution that can be expressed as
\begin{equation}\label{eq:admm3}
\Z^{\ell+1} = (\lambda\widetilde{\X}_t^\top\widetilde{\X}_t+\rho\I)^{-1}
(\lambda\widetilde{\X}_t^\top\widetilde{\X}_t-\Y^{\ell}+\rho\U^{\ell}).
\end{equation}
Note that a naive way to compute matrix inversion in \eqref{eq:admm3} requires $\mathcal{O}(N^3)$ 
arithmetic operations. However, employing the matrix inversion lemma and caching the result 
of the inversion reduces the computational cost to $\mathcal{O}(DN^2)$.

\item  $\U^{\ell+1} = \min_{\U^{\ell}}\L_\rho(\U^{\ell},\Z^{\ell+1},\Y^{\ell})$.

Note that the update of $\U$ also has a closed-form solution given by
\begin{equation}
\begin{aligned}
	&\mathbf{J} = \mathcal{T}_{\frac{1}{\rho}}(\Z^{\ell+1}+\frac{\Y^{\ell}}{\rho}),\\
	&\U^{\ell+1} = \mathbf{J} - \mathrm{diag}(\mathbf{J}),
	\end{aligned}
\end{equation}
where $\mathcal{T}_{\eta}(x) = (|x|-\eta)_+\mathrm{sgn}(x)$ is the so-called 
shrinkage-thresholding operator that acts on each element of the given matrix.

\item  $\Y^{\ell+1} = \Y^{\ell}+\rho(\Z^{\ell+1}-\U^{\ell+1})$, which is a dual gradient ascent 
update with step size $\rho$.
\end{itemize}

The above three steps are repeated until convergence criteria are met or the number of 
iterations exceeds a predefined maximum number. Although here we focus on ADMM as 
the optimization method, similar update rules can be obtained by using more advanced 
techniques including fast and linearized ADMM 
\cite{yang2011alternating,yang2013linearized,goldstein2014fast,lin2015linearized}.

\subsubsection{OMP and OLS-based representation learning}
In each iteration of the OMP and OLS-based representation learning methods, one performs
search over $\mathcal{O}(N)$ data points to identify which among them contribute to the 
innovation representation. In the case of large-scale datasets containing many data points, 
having $\mathcal{O}(N)$ ``oracle calls'' might be prohibitive. Recently, it was shown in 
\cite{mirzasoleiman2014lazier} that when optimizing a submodular function 
\cite{nemhauser1978analysis}, use of a randomized greedy algorithm enables reduction of the 
number of oracle calls to $\mathcal{O}(\frac{N}{k})$ at the cost of a negligible performance 
degradation. While the objective function \eqref{eq:cesmu2} is not submodular, it is {\it weak 
submodular} \cite{das2011submodular,ma}. That is, if the matrix $\X_t^\top\X_t$ is 
well-conditioned (i.e., characterized with a small condition number), the objective function 
\eqref{eq:cesmu2} is close to being submodular. The results of \cite{mirzasoleiman2014lazier}
imply one can still use the randomized greedy method in OMP and OLS instead of the
conventional greedy strategy to accelerate the representation learning process.

The complexity of the OLS-based method can further be reduced using the accelerated OLS 
(AOLS) algorithm, introduced in \cite{hashemi2016sampling}. AOLS improves performance of
 OLS while requiring significantly lower computational costs. As opposed to  OLS which greedily 
 selects data points according to \eqref{eq:aesmuols}, AOLS efficiently builds a collection of 
 orthogonal vectors to represent the basis of $\P^\bot_{\mathcal{A}_{\ell-1}}$ in order to 
 reduce the cost of projection involved in \eqref{eq:aesmuols}. In addition, AOLS anticipates 
 future selections via choosing $L$ data points in each iteration, where $L\geq 1$ is an adjustable 
 hyper-parameter. Selecting multiple data points in each iteration essentially reduces the number 
 of iterations required to identify the representation of data points while typically leading to 
 improved performance. Therefore, in our implementations, we employ the AOLS  strategy instead 
 of OLS to learn the innovation matrix $\U$.
\subsection{Dealing with outliers and missing entries}
The evolving data may contain outliers or missing entries at some or all of the time steps. 
The proposed framework allows for application of convex relaxation methods to handle 
such cases. Specifically, let $\E$ denote a sparse matrix containing outliers, and let $\Omega$ 
denote the set of observed entries of the corrupted data $\X_t^c$. Define the operator 
$\mathcal{P}_\Omega: \R^{D \times N}\rightarrow \R^{D \times N}$ as the orthogonal 
projector onto the span of matrices having zero entries on $[D]\times[N]\backslash \Omega$, 
but agreeing with $\X_t^c$ on entries indexed by the set $\Omega$. Prior to employing
greedy representation learning methods, we identify outliers and values of the missing 
entries by solving the convex program 
\begin{equation}
\begin{aligned}
& \underset{\X_t,\E}{\text{min}}
\quad \|\X_t\|_\ast+\lambda_e \|\E\|_1 \\
& \text{s.t.}\hspace{0.5cm}  \mathcal{P}_\Omega(\X_t^c) = \mathcal{P}_\Omega(\X_t), \quad \X_t^c = \X_t+\E.
\end{aligned}
\end{equation}
Then we can apply the CESM framework using any of the greedy representation 
learning methods to process the ``clean" data $\X_t$, ultimately finding the representations and 
clustering results.

In contrast to the greedy representation learning methods, BP-based approach benefits from 
joint representation learning and corruption elimination. That is, within the CESM 
framework, we may solve
\begin{equation}
\begin{aligned}
& \underset{\X_t,\U,\E}{\text{min}}
\quad \|\U\|_1+\frac{\lambda}{2}\|\widetilde{\X}_t - \X_t \U\|_F^2+\lambda_x\|\X_t\|_\ast+\lambda_e \|\E\|_1 \\
& \text{s.t.}\hspace{0.5cm}  \mathcal{P}_\Omega(\X_t^c) = \mathcal{P}_\Omega(\X_t), \quad \X_t^c = \X_t+\E,\\ 
&\hspace{0.9cm} \widetilde{\X}_t=\frac{1}{\alpha}(\X_t-(1-\alpha)\X_t\C_{t-1}), \quad\mathrm{diag}(\U) = \mathbf{0},
\end{aligned}
\end{equation}
to simultaneously learn the innovation, detect the outliers, and complete the missing entries. 
\section{Simulation Results}\label{sec:sim}
We compare performance of the proposed CESM framework to that of static subspace clustering 
schemes and the evolutionary clustering strategy of AFFECT \cite{xu2014adaptive} on synthetic, 
motion segmentation, and ocean water mass datasets. Note that AFFECT in general does not 
exploit the fact that the data points lie on a union of low dimensional subspaces and its default 
choices for affinity matrix are the negative squared Euclidean distance or its exponential form (i.e., 
an RBF kernel). We found that AFFECT performs poorly compared to other schemes (including 
static algorithms) when using default choices of affinity matrices. Hence, in all experiments we 
use the representation learning methods introduced in Section \ref{sec:alg} for CESM as well as 
for AFFECT to ensure a fair assessment of the proposed evolutionary strategy.\footnote{MATLAB implementation of the proposed algorithm in this paper will be made freely available at \url{https://github.com/realabolfazl}.}
\subsection{Synthetic data}
\begin{figure*}[t]
	\centering
	\minipage[t]{1\linewidth}
	\begin{subfigure}[t]{.245\linewidth}
		\includegraphics[width=\textwidth]{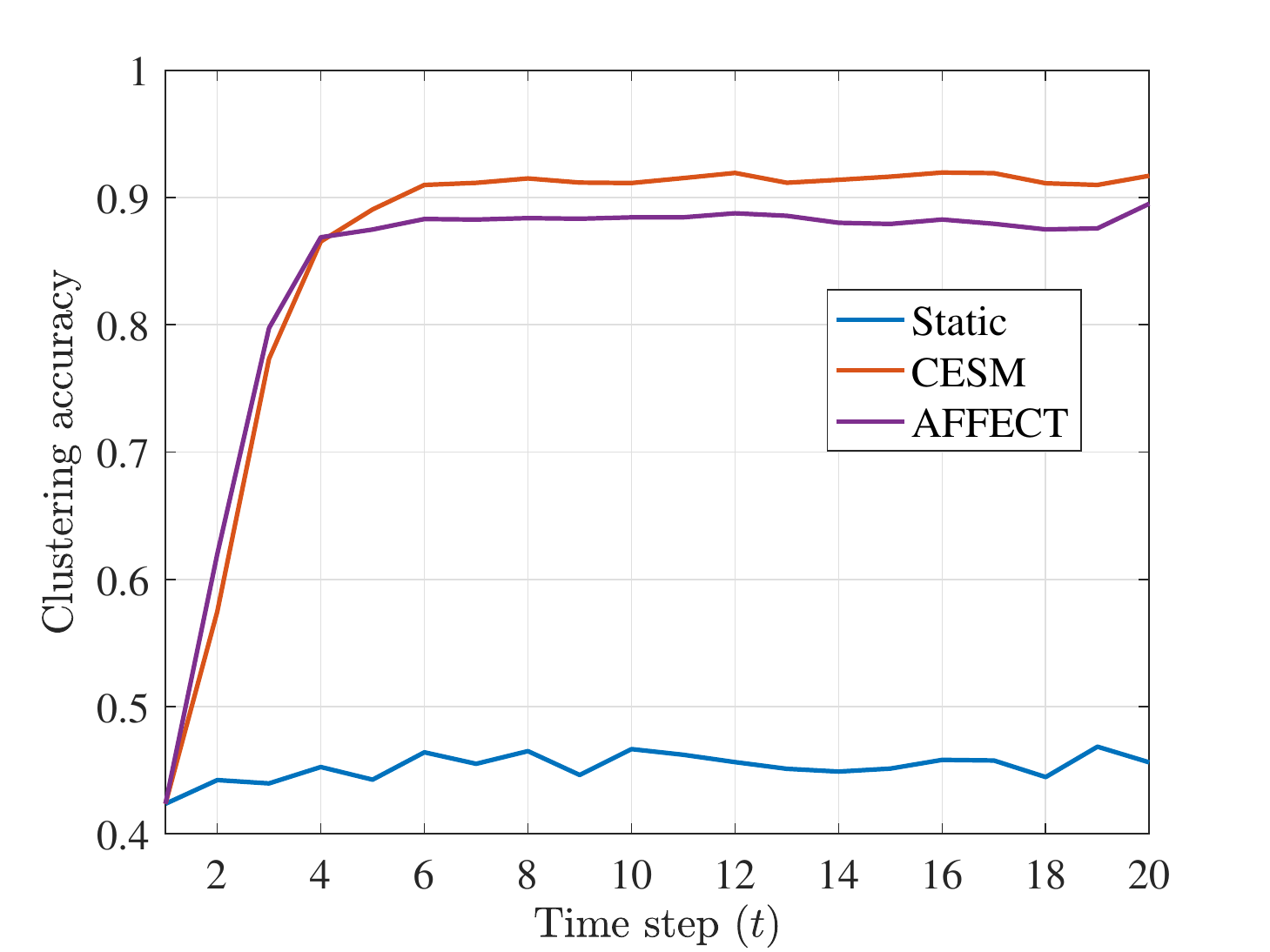}
		\caption{\footnotesize  $45^\circ$ Rotation}
	\end{subfigure}
	\begin{subfigure}[t]{.245\linewidth}
		\includegraphics[width=\linewidth]{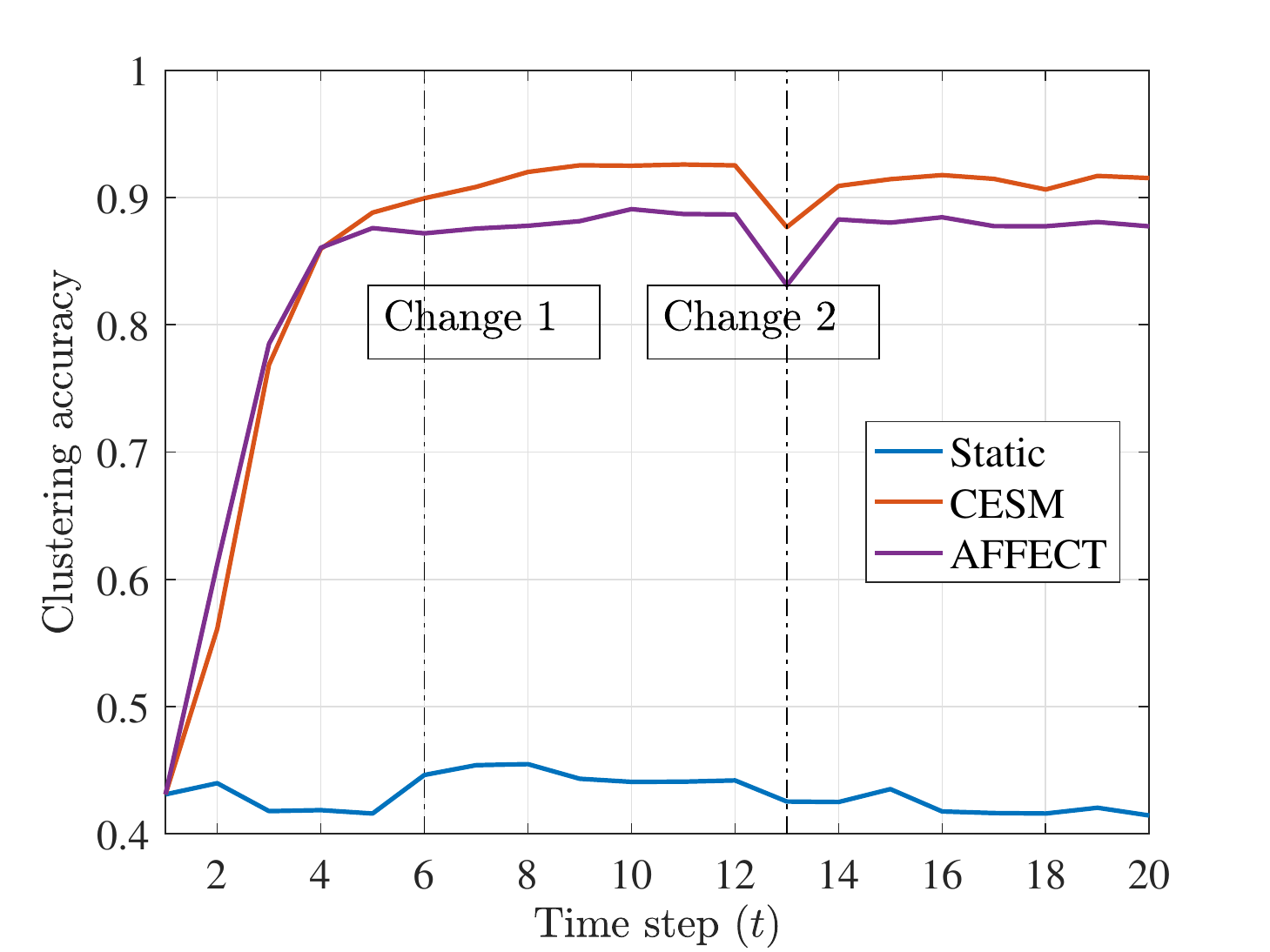}
		\caption{\footnotesize  $45^\circ$ Rotation with subspace change}
	\end{subfigure}
	\begin{subfigure}[t]{.245\linewidth}
	\includegraphics[width=\textwidth]{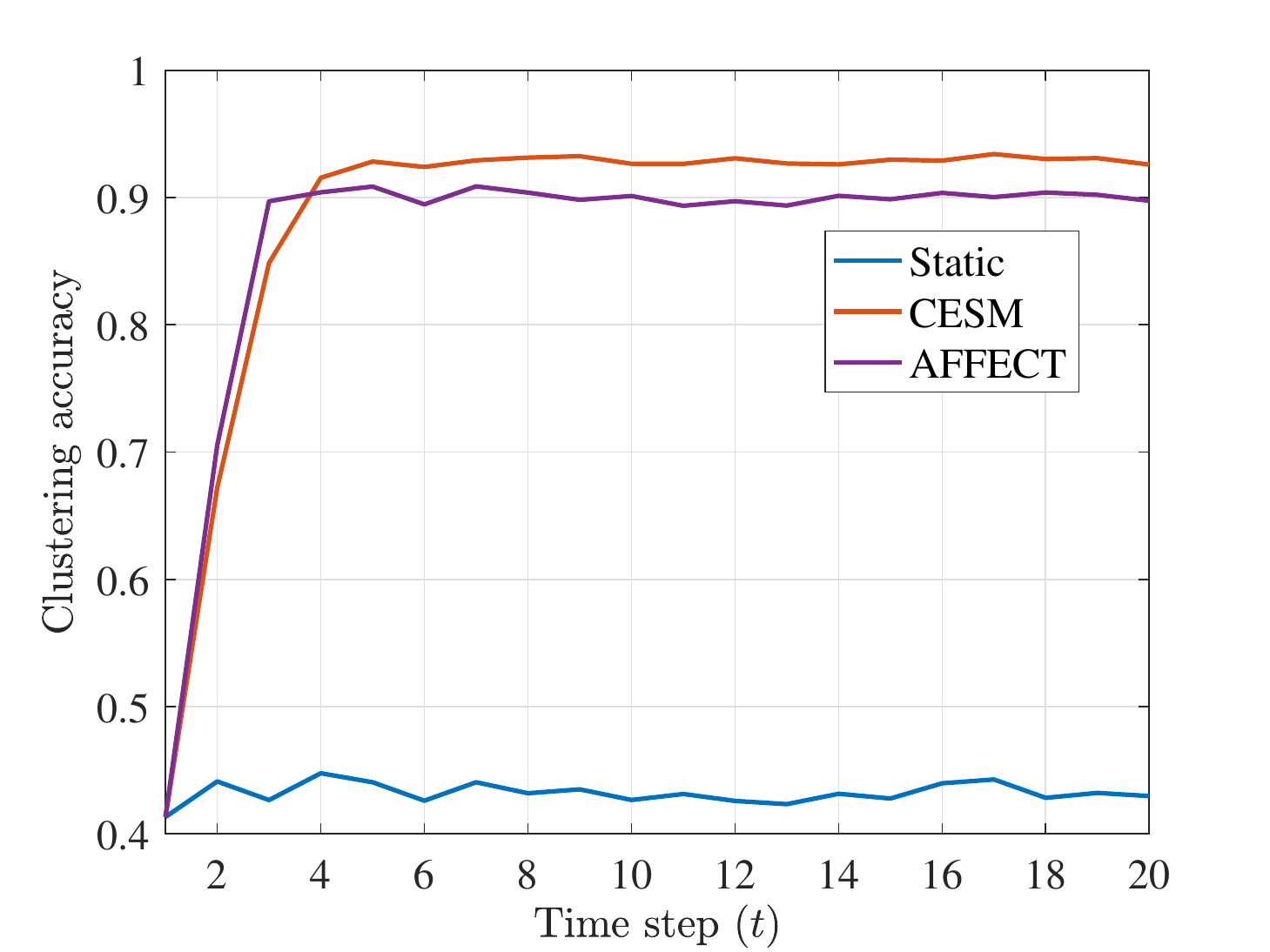}
	\caption{\footnotesize  $90^\circ$ Rotation}
\end{subfigure}
\begin{subfigure}[t]{.245\linewidth}
	\includegraphics[width=\linewidth]{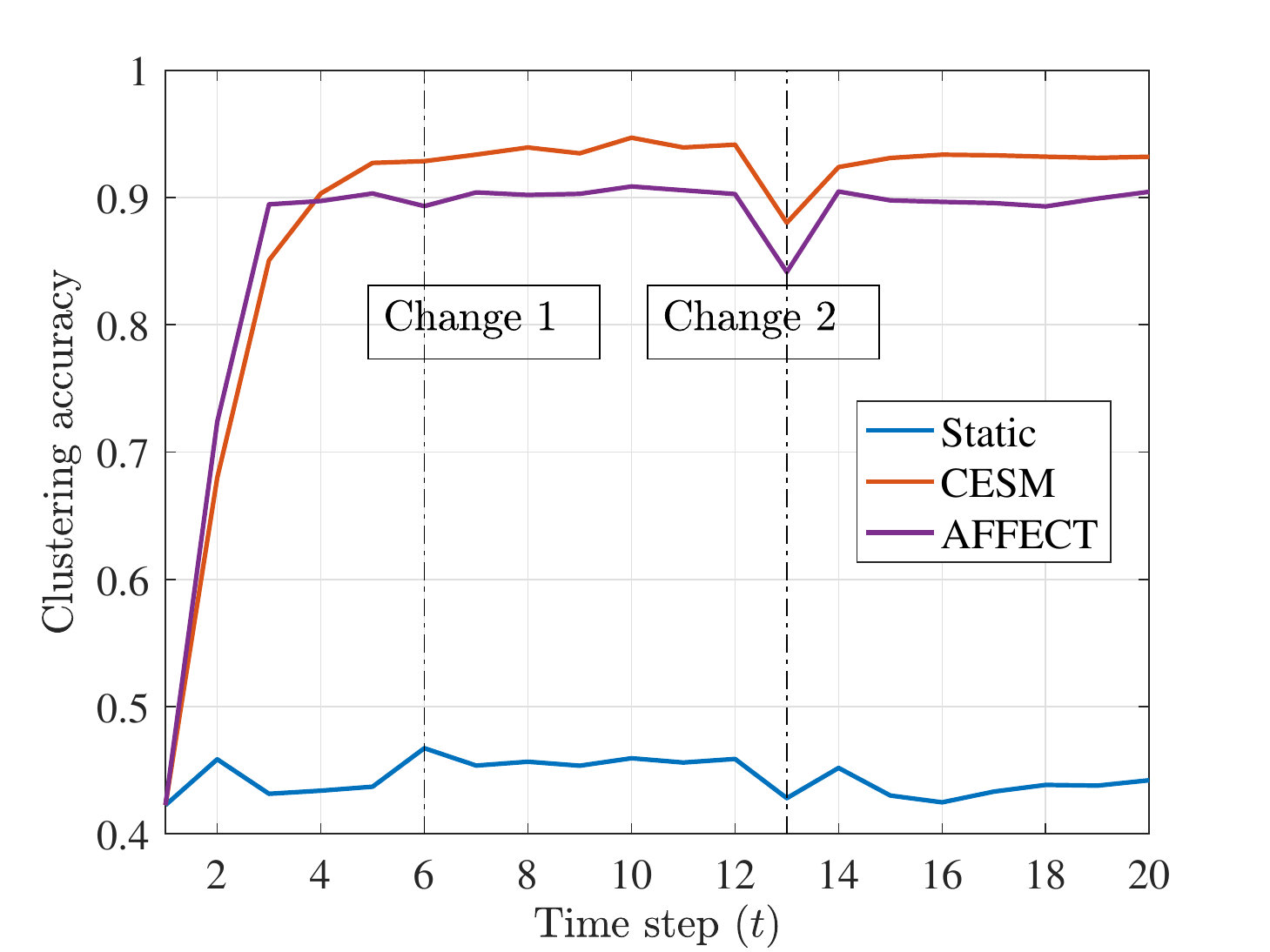}
	\caption{\footnotesize  $90^\circ$ Rotation with subspace change}
\end{subfigure}
	\caption{Comparison of clustering accuracy of static and various evolutionary subspace clustering schemes employing OMP-based representation learning strategy on a simulated data containing 500 points that belong to a union of 10 rotating random subspaces in $\R^{10}$, each of dimension $6$. The proposed CESM framework significantly improves the clustering accuracy and is superior to the AFFECT strategy. Moreover, CESM framework adapts to subspace changes at times $t = 6,13$ as shown in the right-most plots.}
	\label{simacc}
	\endminipage 
	\vspace{-0.3cm}
\end{figure*}
\begin{figure*}[t]
	\centering
	\minipage[t]{1\linewidth}
	\begin{subfigure}[t]{.33\linewidth}
		\includegraphics[width=\textwidth]{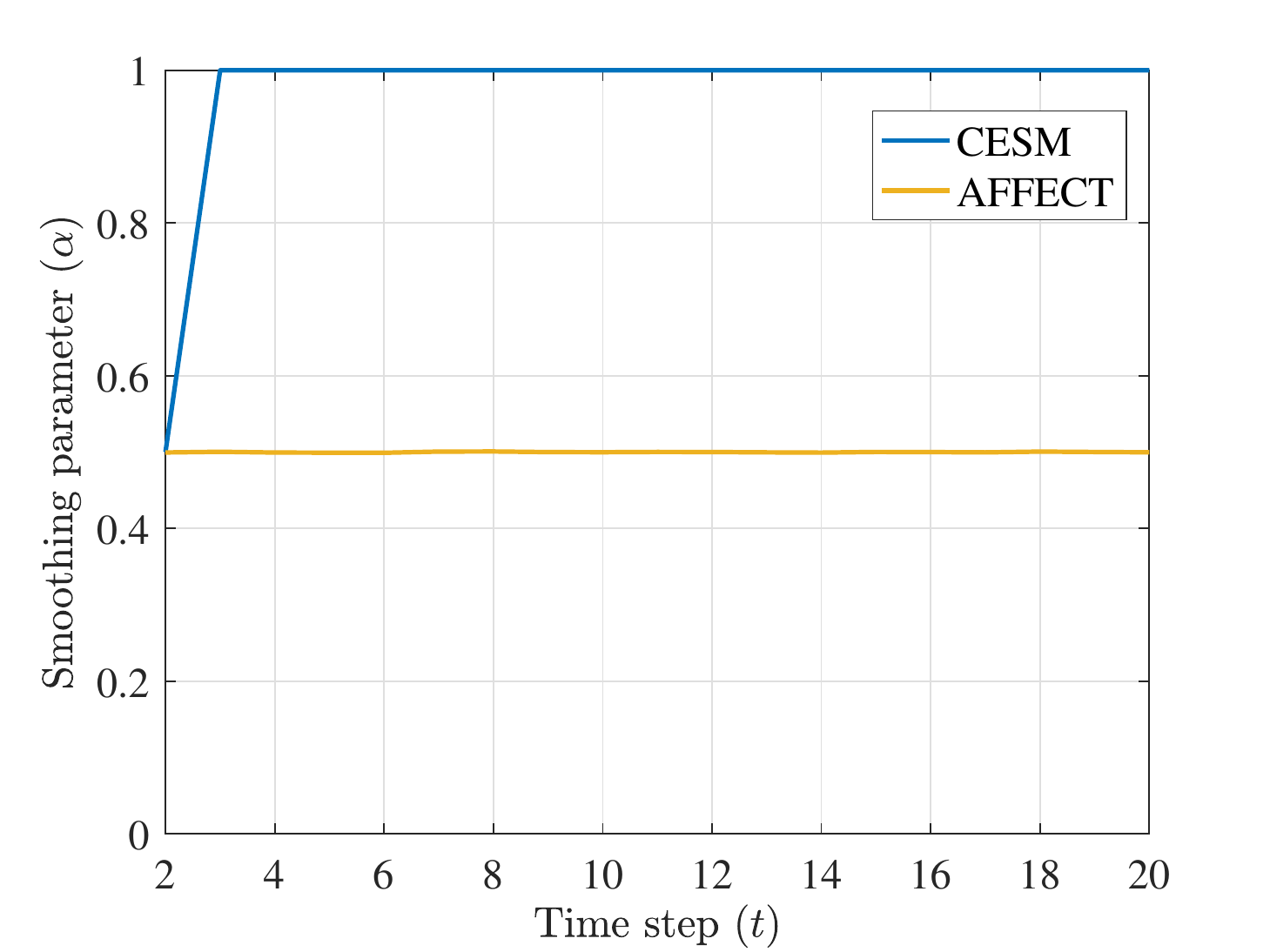}
		\caption{\footnotesize  Static subspaces}
	\end{subfigure}
	\begin{subfigure}[t]{.33\linewidth}
		\includegraphics[width=\linewidth]{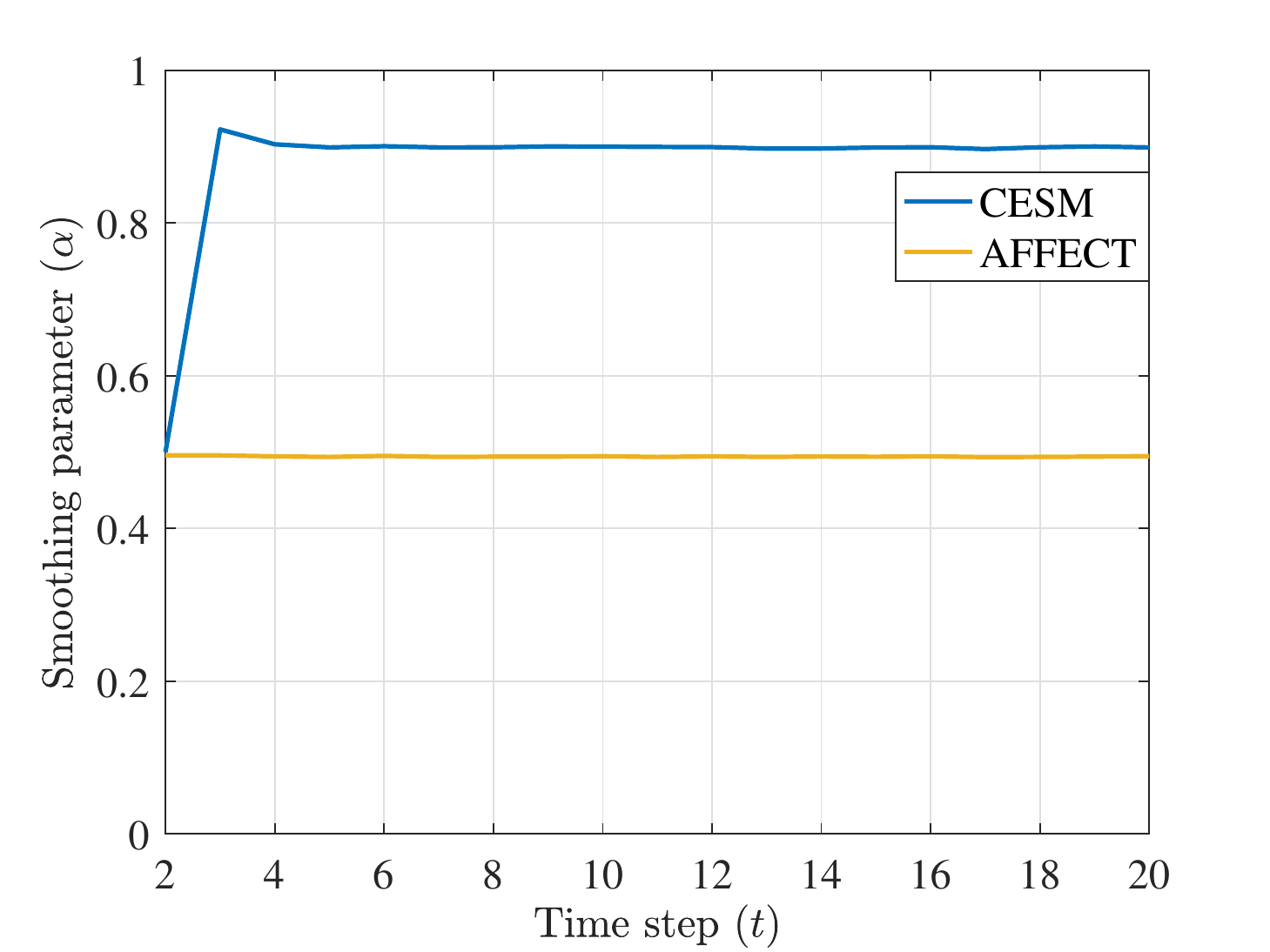}
		\caption{\footnotesize $45^\circ$ Rotation}
	\end{subfigure}
	\begin{subfigure}[t]{.33\linewidth}
		\includegraphics[width=\linewidth]{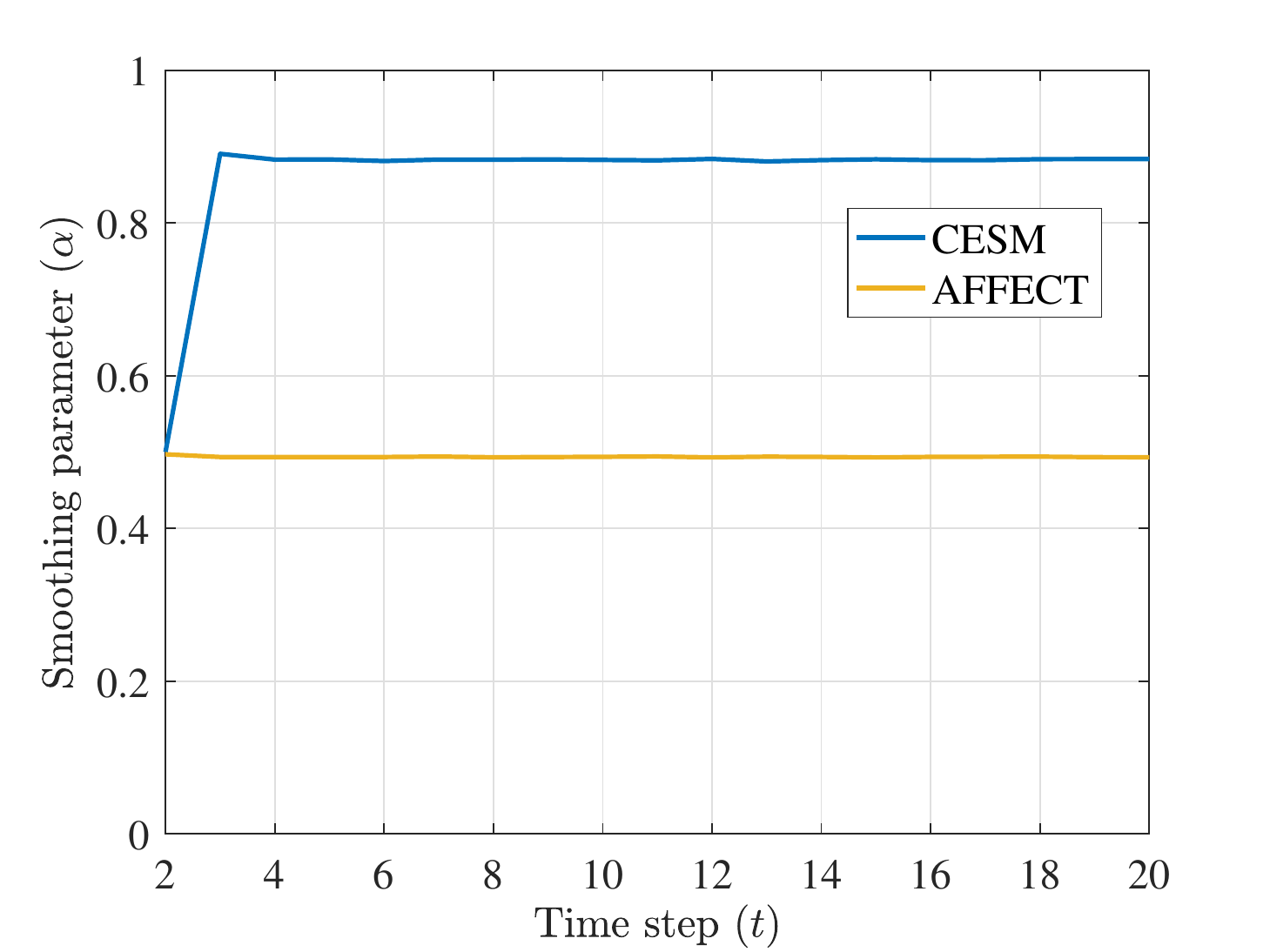}
		\caption{\footnotesize $90^\circ$ Rotation}
	\end{subfigure}
	\begin{subfigure}[t]{.33\linewidth}
	\includegraphics[width=\textwidth]{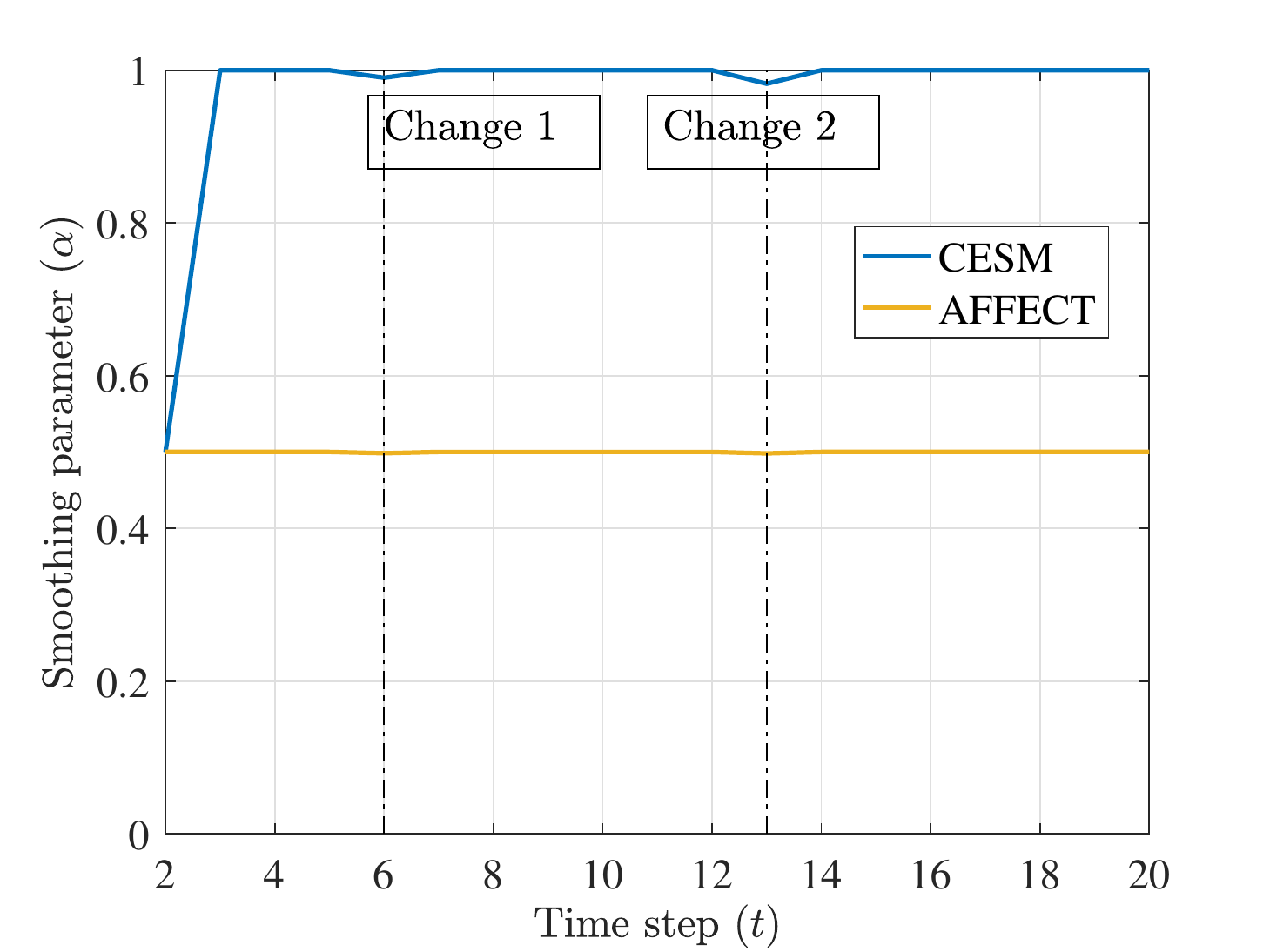}
	\caption{\footnotesize  Static subspaces with subspace change}
\end{subfigure}
\begin{subfigure}[t]{.33\linewidth}
	\includegraphics[width=\linewidth]{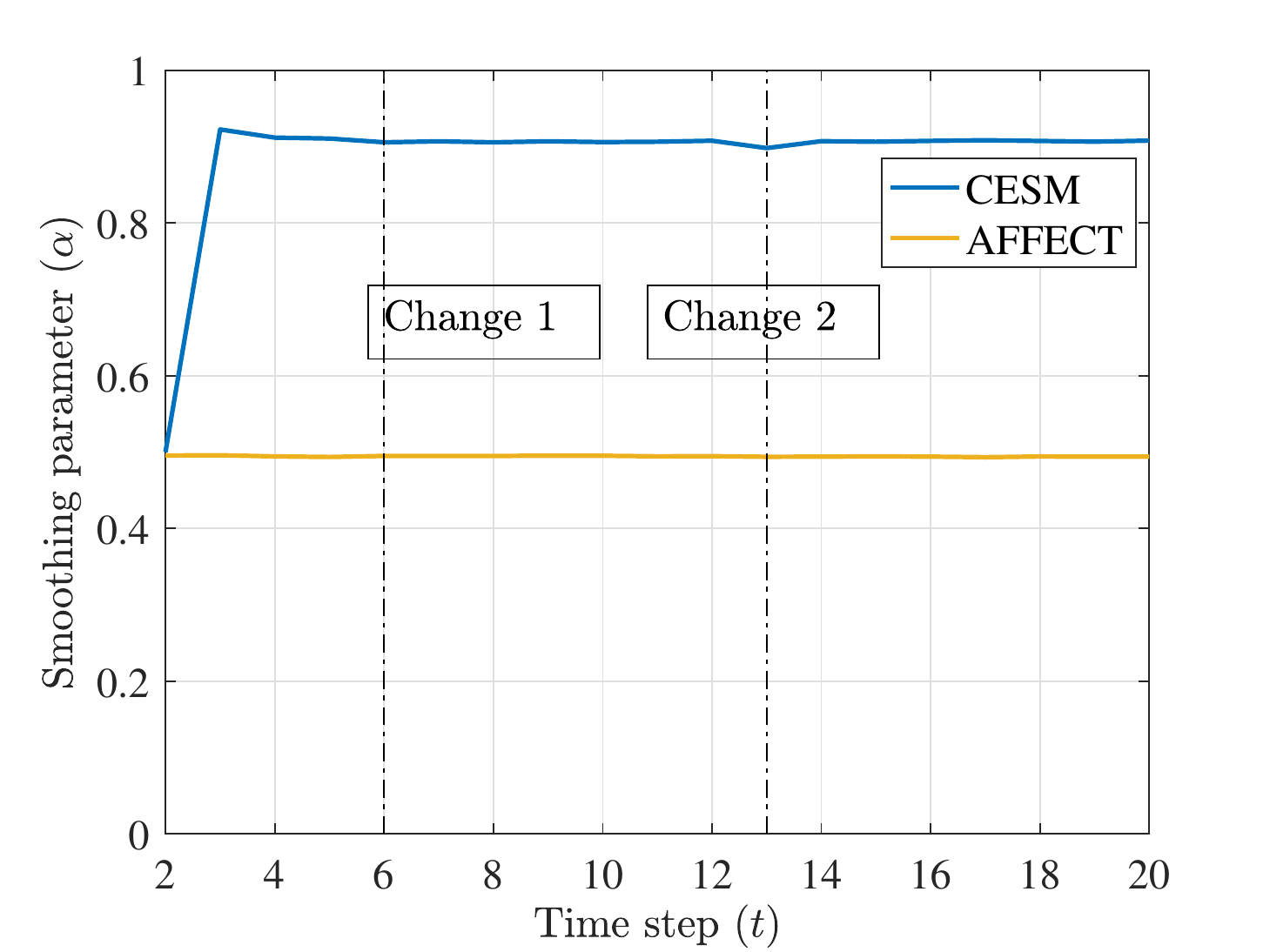}
	\caption{\footnotesize $45^\circ$ Rotation  with subspace change}
\end{subfigure}
\begin{subfigure}[t]{.33\linewidth}
	\includegraphics[width=\linewidth]{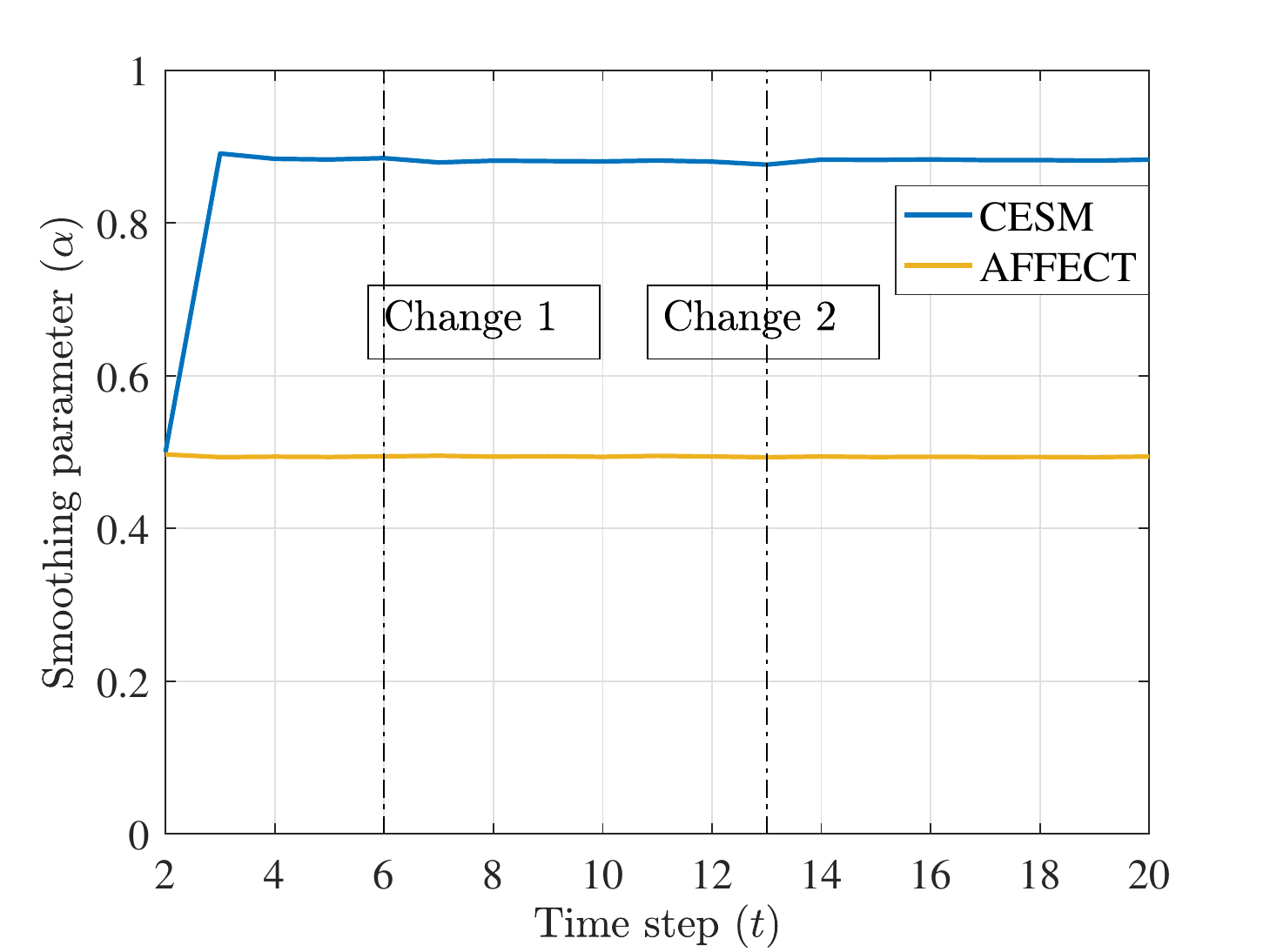}
	\caption{\footnotesize $90^\circ$ Rotation with subspace change}
\end{subfigure}
	\caption{Comparison of the smoothing parameter $\alpha$ for various 
	evolutionary subspace clustering schemes employing OMP-based 
	representation learning strategy on a simulated data containing 500 
	points lying on a union of 10 rotating random subspaces in $\R^{10}$, 
	each of dimension $6$. AFFECT's smoothing parameter remains
	approximately constant regardless of the underlying evolutionary behavior 
	while the smoothing parameter for the CESM framework 
	dynamically reflects the structure and reacts to cluster changes.}
	\label{simalpha}
	\endminipage 
	\vspace{-0.3cm}
\end{figure*}
\begin{figure*}[t]
	\minipage[t]{1\linewidth}
	\begin{subfigure}[t]{.245\linewidth}
		\includegraphics[width=\textwidth]{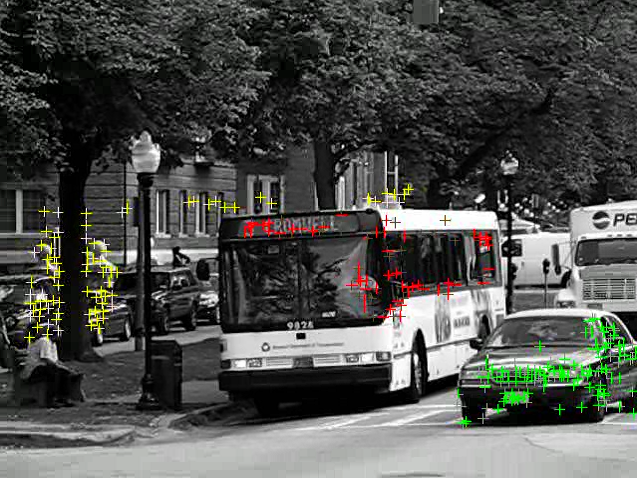}
	\end{subfigure}
	\begin{subfigure}[t]{.245\linewidth}
		\includegraphics[width=\linewidth]{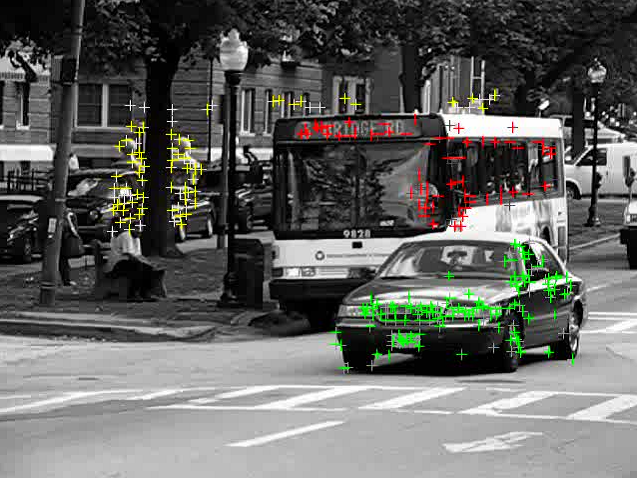}
	\end{subfigure}
	\begin{subfigure}[t]{.245\linewidth}
		\includegraphics[width=\linewidth]{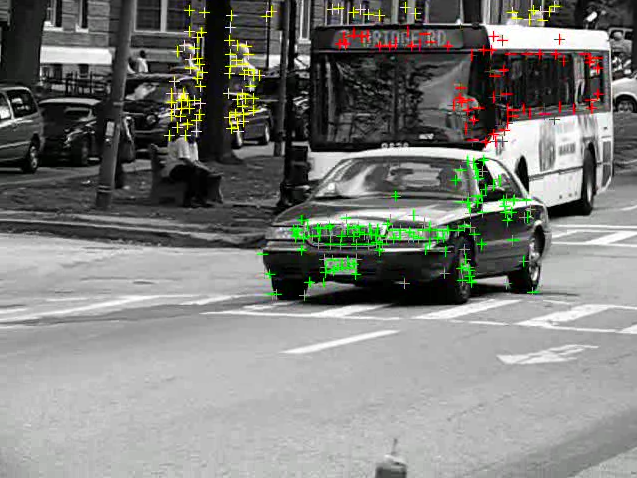}
	\end{subfigure}
	\begin{subfigure}[t]{.245\linewidth}
		\includegraphics[width=\linewidth]{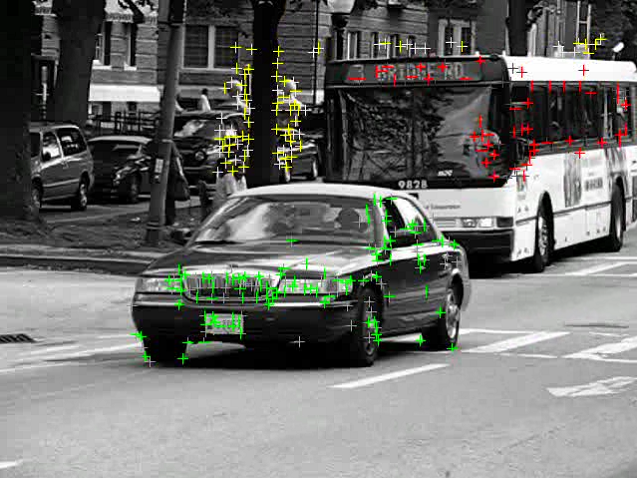}
	\end{subfigure}
	\caption{Example frames from the videos in the Hopkins 155 dataset \cite{tron2007benchmark}.}
	\label{motionseg}
	\endminipage 
	\vspace{-0.3cm}
\end{figure*}
In a variety of applications including motion segmentation \cite{tron2007benchmark}, 
the data points and their corresponding subspaces are characterized by rotational 
and transitional motions. Therefore, to simulate an underlying evolutionary process 
for data points lying on a union of subspaces, we consider the following scenario of 
rotating subspaces  where we repeat each experiment for 150 trials.

At time $ t = 1$, we construct $n = 10$ linear subspaces in $\R^D$, $D = 10$, each 
with  dimension $d = 6$ by choosing their bases as the top $d$ left singular vectors 
of a random Gaussian matrix in $\R^{D\times D}$. Then, we sample $N = 500$ data 
points, 50 from each subspace, by projecting random Gaussian vectors to the span 
of each subspace. Note that, in this setting, all the subspaces are distributed uniformly 
at random in the ambient space and all data points are uniformly distributed on the 
unit sphere of each subspace. According to the analysis in 
\cite{soltanolkotabi2012geometric,soltanolkotabi2014robust,you2015sparse}, this in 
turn implies that the subspace preserving property and the performance of representation 
learning methods based on BP, OMP, and AOLS is similar. However, we intentionally 
generate relatively low number of data points compared to the dimension of 
subspaces and the dimension of the ambient space; this creates a setting that is
challenging for static subspace clustering algorithms. After constructing subspaces 
at time $t = 1$, we evolve the subspaces by rotating their basis $45^\circ$ or 
$90^\circ$ around a random vector and project the data points $\X_1$ on the span 
of the rotated subspaces to obtain $\X_2$. We continue this process for 
$T = 20$ time steps. Note that for each subspace we perform rotation around a 
different random vector. Otherwise, if the rotations were around the same vector, the 
above setting would not be an evolutionary process as the relative positions of 
subspaces and data points would not vary over time. 
For brevity, we only present results of using OMP-based learning to 
find the representation matrices for static and competing evolutionary subspace 
clustering algorithms; however, we observed similar results for representation 
learning methods based on BP and AOLS. 

Next, we consider a related experiment where in addition to rotation, at time $t = 6$ 
all data points generated from subspace $\S_{10}$ are absorbed by subspace
$\S_9$. That is, at $t = 6$ we project $\X_{5,{\S_{10}}}$ to the span of $\S_9$. At 
time $t=13$, these data points are separated from $\S_9$ and lie once again on 
$\S_{10}$. Hence, for $6\leq t\leq 12$ the effective number of subspaces is $n=9$ 
and there are 100 data points in $\S_9$. 

The clustering accuracy results for these two experiments are illustrated in Fig. 
\oldref{simacc}. For the first experiment, as seen from 
Fig. \oldref{simacc} (a) and Fig. \oldref{simacc} (c), the static SSC-OMP algorithm 
performs poorly compared to  CESM and AFFECT. Since CESM 
and AFFECT exploit the evolutionary behavior of the data points, after a few time 
steps their accuracy significantly increases. We further observe that the proposed 
CESM framework achieves better accuracy than AFFECT; this is likely because the former
exploits the self-expressiveness property of data points in the representation 
learning process while the latter simply combines current and prior representations to 
enforce the self-expressiveness property.

A comparison of the performance results in the second experiment is shown in 
Fig. \oldref{simacc} (b) and Fig. \oldref{simacc} (d). We observe that the 
performance of all evolutionary schemes suffers temporary degradations at times 
$t = 6$ and $t = 13$. The reason for this phenomenon is that the data points $\X_{6,{\S_{10}}}$ at $t=6$ 
are significantly different from $\X_{5,{\S_{10}}}$ at time $t=5$ due to being absorbed by $\S_9$ 
at time $t = 6$ and not belonging to $\S_{10}$. Therefore, since the subspaces are nearly 
independent, prior representations  $\{\c_{5,i}\}_{i\in \S_{10}}$ and  $\{\c_{12,i}\}_{i\in \S_{10}}$
are simply not well-aligned with the sudden changes taking place at times $t = 6,13$.
We further note that the deterioration in clustering accuracy is more severe for 
AFFECT than for CESM. We also observe from the figure that the proposed 
evolutionary scheme is able to quickly adapt to changes. At $t = 13$, the data 
points that were previously absorbed by $\S_{9}$ are projected back to the span of $\S_{10}$;
as a result of this change, the performance of evolutionary schemes decreases. 
However, accuracy of the evolutionary methods recovers at $t=14$ and improves 
onward as they exploit the evolving property of the data. Similar to the first experiment, 
due to exploiting the fact that data points lie on a union of subspaces, the proposed 
CESM framework outperforms the AFFECT's strategy.

Next, we investigate the value of $\alpha$, i.e., the smoothing parameter discovered and used 
by CESM and AFFECT in the previously described experiments to further assess which scheme 
more accurately captures the evolutionary nature of the subspaces. Fig. \oldref{simalpha} 
illustrates changes in the value of $\alpha$ over time, where in addition to the above two 
experiments we consider the scenario where subspaces are not rotating. The figure indicates 
that the smoothing parameter of AFFECT remains approximately $0.5$ regardless of how
rapidly the subspaces evolve. Note that the smoothing parameter essentially quantifies 
evolutionary character of a dataset: if the data is static, we expect $\alpha = 0$ or $\alpha = 1$ 
for both CESM and AFFECT. As opposed to the AFFECT's smoothing parameter, the value of 
$\alpha$ for the CESM framework quickly converges to the anticipated level; note that we 
initialized $\alpha$ as $0.5$. Fig. \oldref{simalpha} (d)-(f) further suggest that the smoothing 
parameter of the proposed CESM framework noticeably changes at times $t = 6,13$. This is 
a strong indication that CESM is capable of detecting subspace changes at 
$t = 6,13$, while AFFECT fails to detect that the subspaces are rotating. 

The above results suggest that the proposed framework improves performance of static 
subspace clustering algorithms when the data is evolving, while also being superior to 
state-of-the-art evolutionary clustering strategies in the considered settings. In contrast to 
prior schemes, the smoothing parameter of the proposed framework is meaningful and 
interpretable, and timely adapts to the underlying evolutionary behavior of the subspaces.
\subsection{Real-time motion segmentation}\label{sec:sim-motion}
\begin{table*}[t]\centering
	\caption{Performance comparison of static and various evolutionary subspace clustering algorithms 
	on real-time motion segmentation dataset. The best results for each row are in boldface fonts. For the 
	CESM framework, the top results in each row correspond to the case of using a constant smoothing 
	factor with the lowest average error while the bottom results in each row are achieved by using the 
	proposed alternating minimization schemes to learn the smoothing parameter at each time step.}
	\ra{1}
	\begin{tabular*}{1\linewidth}{@{}ccccccccccccc@{}}\toprule
	&\phantom{}&\multicolumn{3}{c}{Static} &\phantom{}& \multicolumn{3}{c}{AFFECT}&\phantom{}& \multicolumn{3}{c}{CESM} \\
	\cmidrule{3-5} \cmidrule{7-9} \cmidrule{11-13} 
	Learning method &&error (\%)&RI (\%)&runtime (s)&&error (\%)&RI (\%)&runtime (s)&&error (\%)&RI (\%)&runtime (s)\\ \midrule
	BP            &&10.76& 86.29 &46.16    &&  9.86 &  87.78 &47.35 &&\begin{tabular}{@{}c@{}} 8.88\\ \bf8.77 \end{tabular}&\begin{tabular}{@{}c@{}}  \bf89.33\\ 89.14  \end{tabular} &\begin{tabular}{@{}c@{}} 45.10\\ \bf 41.21 \end{tabular}   \\\midrule
	OMP           &&31.66&  62.00 & 1.80   &&14.47 &  86.21  &3.31 &&\begin{tabular}{@{}c@{}}\bf 5.54 \\6.85 \end{tabular} &\begin{tabular}{@{}c@{}} \bf 93.25\\ 88.23  \end{tabular}   &\begin{tabular}{@{}c@{}}  \bf 0.90\\0.93 \end{tabular}   \\\midrule
	AOLS ($L = 1$) &&16.27 & 78.41 & 4.08   &&9.27 &  90.76 & 5.39&&\begin{tabular}{@{}c@{}}\bf6.57 \\8.24 \end{tabular}    &\begin{tabular}{@{}c@{}} \bf 91.97\\ 90.12  \end{tabular}&\begin{tabular}{@{}c@{}}2.07\\\bf1.93 \end{tabular}   \\\midrule
	AOLS ($L = 2$) &&8.54& 89.10  & 3.75   && 6.17&  93.08 &5.17 &&\begin{tabular}{@{}c@{}}\bf5.25\\5.70 \end{tabular}     &\begin{tabular}{@{}c@{}} \bf93.35\\ 92.85  \end{tabular}& \begin{tabular}{@{}c@{}}1.85\\\bf1.77 \end{tabular}  \\\midrule
	AOLS ($L = 3$) &&6.97 & 91.09 & 3.14   && 5.92 &  93.40 & 4.28&& \begin{tabular}{@{}c@{}}\bf 5.49\\5.60 \end{tabular}    &\begin{tabular}{@{}c@{}} \bf 94.17\\ 93.90  \end{tabular}&\begin{tabular}{@{}c@{}}1.70\\\bf1.69 \end{tabular}   \\			
	\bottomrule
\end{tabular*} 
	\label{hopkins}
	\vspace{-0.3cm}
\end{table*}
Motion segmentation is the problem of clustering a set of two dimensional trajectories 
extracted from a video sequence with multiple rigidly moving objects into groups; the 
resulting clusters correspond to different spatiotemporal regions (Fig. \oldref{motionseg}). 
The video sequence is often received as a stream of frames and it is desirable 
to perform motion segmentation in a real-time fashion \cite{smith1995asset,collins2005online}. 
In the real-time setting, the $t\ts{th}$ snapshot of $\X_t$ (a time interval consisting of multiple 
video frames) is of dimension $2F_t\times N_t$, where $N_t$ is the number of trajectories 
at $t\ts{th}$ time interval, $F_t$ is the number of video frames received in $t\ts{th}$ time interval, $n_t$ is 
the number of rigid motions at $t\ts{th}$ time interval, and $F = \sum_t F_t$ denotes the 
total number of frames. Real-time motion segmentation falls within the scope of
evolutionary subspace clustering since the received video sequence is naturally characterized 
by temporal properties; at $t\ts{th}$ time interval, the trajectories of $n_t$ rigid 
motions lie in a union of $n_t$ low-dimensional subspaces in $\R^{2F_t}$, each 
with the dimension of at most $d_t = 3n_t$ \cite{tomasi1992shape}. 

In contrast to the real-time motion segmentation, clustering in offline settings
is performed on the entire sequence, i.e., $\X = {[{\X}_1^\top,\dots,{\X}_T^\top]}^\top$. 
Therefore, one expects to achieve more accurate segmentation in the offline settings.
However, offline motion segmentation cannot be used in scenarios where 
some motions vanish or new motions appear in the video, or in cases where a 
real-time motion segmentation solution is desired.

To benchmark the performance of the proposed CESM framework, we 
consider Hopkins 155 database \cite{tron2007benchmark} which consists of 155 video 
sequences with 2 or 3 motions in each video (corresponding to 2 or 3 low-dimensional 
subspaces). Unlike the majority of prior work that process this data set in an 
offline setting, we consider the following real-time scenario: each video is divided into 
$T$ data matrices $\{\X_t\}_{t=1}^T$ such that $F_t \geq 2n$ for a video with $n$ 
motions.  Then, we apply PCA on $\X_t$ and take its top $D = 4n$ left singular vectors 
as the final input to the representation learning algorithms.

We benchmark the proposed framework by comparing it to static subspace 
clustering and AFFECT; the former applies subspace clustering at each time step 
independently from the previous clustering results while the latter applies spectral 
clustering \cite{ng2001spectral} on the weighted average of affinity matrices $\A_t$ and 
$\A_{t-1}$. The default choices for the affinity matrix in AFFECT are the negative 
squared Euclidean distance or its exponential form. Under these choices, AFFECT 
achieves a clustering error of  $44.1542$ and $21.9643$ percent for the negative 
squared Euclidean distance or its exponential form, respectively, which as we 
present next is inferior even to the static subspace clustering algorithms. Hence, 
to fairly compare the performance of different evolutionary clustering strategies, 
we employ BP \cite{chen2001atomic,elhamifar2009sparse,elhamifar2013sparse}, 
OMP \cite{pati1993orthogonal,dyer2013greedy,you2015sparse}, and AOLS 
\cite{hashemi2016sparse,hashemi2017accelerated} with $L = 1,2,3$ to learn the 
representations for all schemes, including AFFECT.

The performance of various schemes are presented in 
Table \oldref{hopkins};
 there, the results are averaged over all sequences and all 
 time intervals excluding the initial time interval $t = 1$. The initial time interval is excluded
because for a specific representation learning method (e.g., BP), the results of 
static subspace clustering and evolutionary schemes coincide. Note that for the 
proposed CESM framework, the top results in each row of Table \oldref{hopkins} 
correspond to the case of using a constant smoothing factor with 
the lowest average error while the bottom results in each row are achieved by using 
the proposed alternating minimization schemes to learn the best smoothing parameter 
for each time interval.

As we can see from the table, static subspace clustering has higher clustering 
errors than their evolutionary counterparts; this is due to not incorporating any 
knowledge about the representations of the data points at other times. Furthermore, 
the proposed CESM framework is superior to AFFECT in terms 
of clustering error for all the representation learning methods. In 
addition, the proposed CESM framework achieves lower running 
time than static and AFFECT strategies, especially for the case of using OMP 
and AOLS as the representation learning methods. This supports the observation 
that CESM promotes sparser $\U_t$ by leveraging $\C_{t-1}$ in the 
process of learning $\C_t$ which in turn leads to faster convergence of OMP 
and AOLS. Similar to what we observed on synthetic datasets, the smoothing 
parameter of AFFECT (with both the default choices for the affinity matrix and 
the SSC-based affinity learning methods) was 
approximately $0.5$ for all sequences and thus unable to capture evolutionary 
structure of the subspaces in a meaningful and interpretable manner.
 
\subsection{Ocean water mass clustering}
\begin{figure*}[t]
	\centering
	\minipage[t]{1\linewidth}
	\begin{subfigure}[t]{.33\linewidth}
		\includegraphics[width=\textwidth]{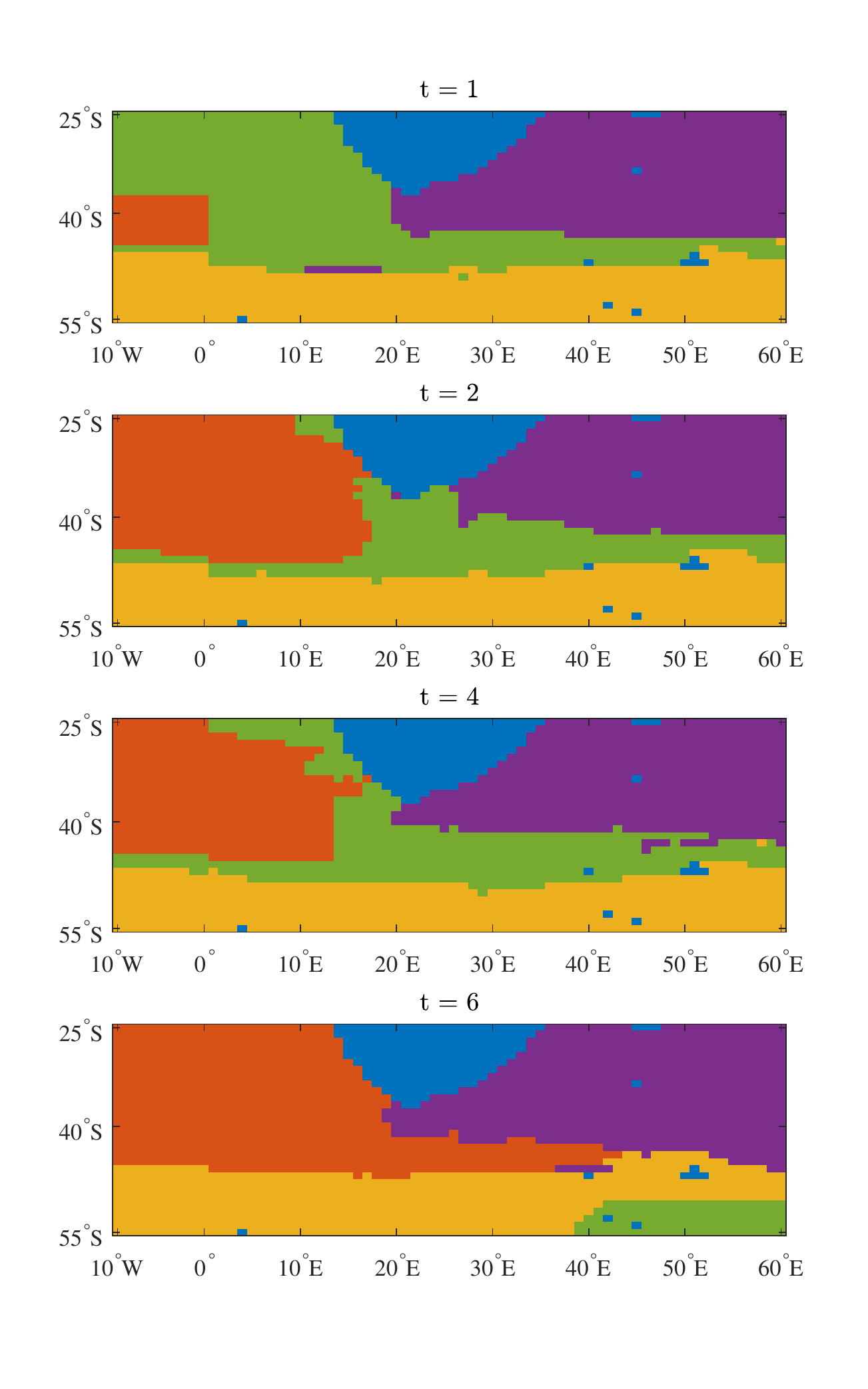}
		\vspace{-1cm}
		\caption{\footnotesize  static}
	\end{subfigure}
	\begin{subfigure}[t]{.33\linewidth}
		\includegraphics[width=\linewidth]{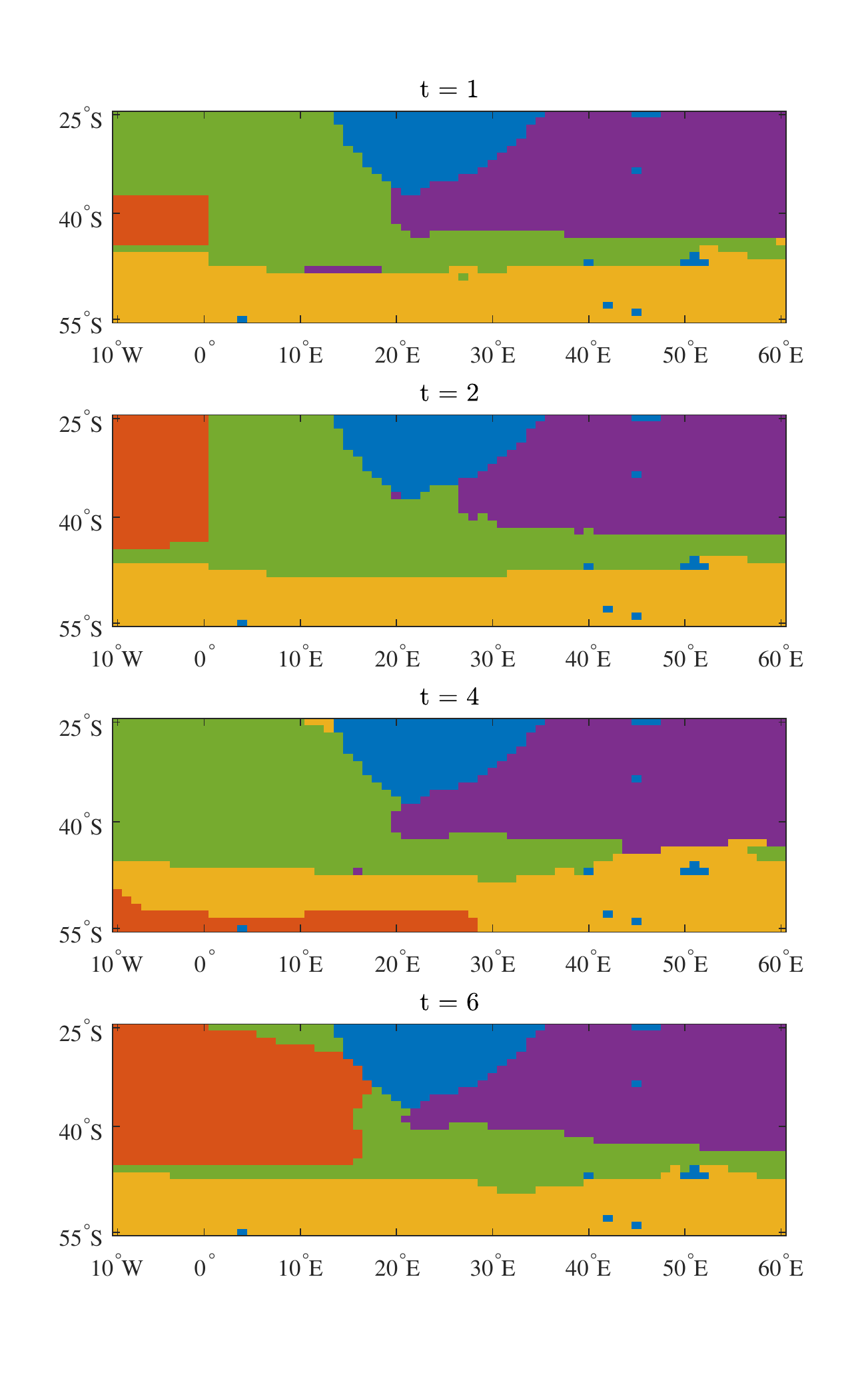}
		\vspace{-1cm}
		\caption{\footnotesize AFFECT}
	\end{subfigure}
	\begin{subfigure}[t]{.33\linewidth}
	\includegraphics[width=\linewidth]{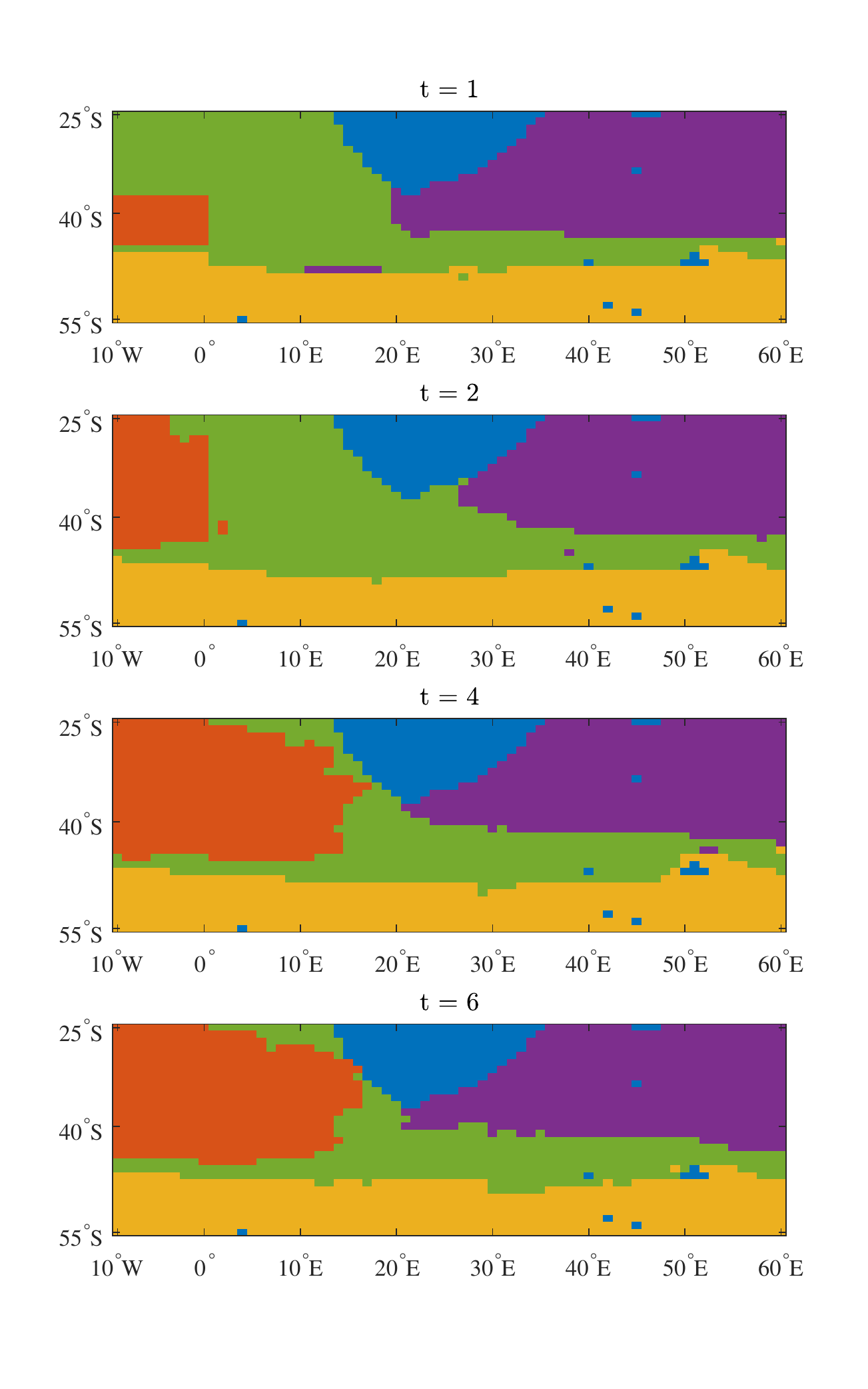}
	\vspace{-1cm}
	\caption{\footnotesize CESM}
\end{subfigure}
	\caption{Clustering results of {\color{black}four different types of water masses at 1000 dbar near the coast of south Africa (colored with blue)}. using static and various evolutionary subspace clustering schemes employing AOLS-based representation learning strategy with $L = 3$. The static subspace clustering scheme and AFFECT 
fail to keep track of the orange water mass at time $t = 6$ and $t=4$, respectively. However, our proposed CESM framework detects homogeneous water
masses across all time steps.}
	\label{ocean}
	\endminipage 
	\vspace{-0.3cm}
\end{figure*}
\begin{table}[t]\centering
	\caption{Average salinity and temperature of {\color{black}four different types of water masses at 1000 dbar near the coast of south Africa} identified by CESM framework employing AOLS-based representation learning strategy with $L = 3$ at different time steps. The results in top, middle and bottom for each cluster correspond to $t = 2,4,6$, respectively.}
	\ra{1}
	\begin{tabular*}{0.5\linewidth}{@{}ccccccccccccc@{}}\toprule
		water mass&\phantom{11111}&salinity level&\phantom{11111}&temperature ($^\circ$C) \\
		\midrule
		 orange
		 &&\begin{tabular}{@{}c@{}}34.4554 \\34.3564  \\ 34.5008\end{tabular}    
		 && \begin{tabular}{@{}c@{}} 3.4971\\3.6164 \\3.2141 \end{tabular} \\\midrule	
		 green
	     &&\begin{tabular}{@{}c@{}}34.3452  \\34.6693 \\34.3640 \end{tabular}   
	      && \begin{tabular}{@{}c@{}}3.5849 \\1.9910 \\3.6482 \end{tabular} \\\midrule	
		 yellow 
		    &&\begin{tabular}{@{}c@{}}34.6603 \\34.4974 \\34.6680 \end{tabular}    
		    && \begin{tabular}{@{}c@{}} 2.0177\\6.4445 \\2.0914 \end{tabular} \\\midrule	
		 purple
		  &&\begin{tabular}{@{}c@{}}34.4998 \\34.4649 \\34.4997 \end{tabular}    
		  && \begin{tabular}{@{}c@{}}6.3313 \\3.4162 \\6.5522 \end{tabular} \\		
		\bottomrule
	\end{tabular*} 
	\label{oceant}
	\vspace{-0.3cm}
\end{table}
Ocean temperature and salinity has been tracked by Argo ocean observatory 
system comprising more than 3000 floats which provide 100,000 plus temperature 
and salinity profiles each year. These floats cycle between the ocean surface and 
2000m depth every 10 days, taking salinity and temperature measurements at 
varying depths. A water mass is characterized as a body of water with a common 
formation and homogeneous features, such as salinity and temperature. Study of 
water masses can provide insight into climate change, seasonal climatological 
variations, ocean biogeochemistry, and ocean circulation and its effect on transport 
of oxygen and organisms, which in turn affects the biological diversity of an area. 

To illustrate the abilities of evolutionary subspace clustering in modeling various 
real-world problems, including those outside the computer vision community, we 
analyze the global gridded dataset produced via the Barnes method that was 
collected and made freely available by the international Argo program. This dataset  
contains monthly averages (since January 2004) of ocean temperature and salinity
with 1 degree resolution worldwide \cite{roemmich20092004,li2017development}. 

In order to identify homogeneous water masses, we apply static and various 
evolutionary subspace clustering schemes, using AOLS-based representation 
learning method with $L = 3$ on the temperature and salinity data at the location 
near the coast of South Africa where the Indian Ocean meets the South Atlantic 
(specifically, the area located at latitudes 25$^\circ$ S to 55$^\circ$ S and 
longitudes 10$^\circ$ W to 60$^\circ$ E). 

According to prior studies in \cite{arzeno2016outcome,pickard2016descriptive,arzeno2017evol}, 
there are three well-known and strong mater masses in this area: (1) Agulhas currents,
(2) the Antarctic intermediate water (AAIW), 
and (3) the circumpolar deep water mass.
Therefore, following the discussion in \cite{arzeno2016outcome} we set 
the number of clusters to $n = 4$ to further account for other water masses in 
the area.

The area described above accounts for $N = 1921$ evolving data points, each 
containing the monthly salinity and temperature from April to September for two 
years acquired starting in the year 2004 and 2005 ($t = 1$) until year 2014 and 
2015 ($t = 6$). Temperature and salinity were normalized by subtracting the 
mean and dividing by the standard deviation of the entire time frame of interest. 
This procedure results in $24\times N$ data matrices $\{\X_t\}_{t=1}^6$ which 
are then used as inputs to the evolutionary subspace clustering algorithms. As 
stated in Section \oldref{sec:ext}, we employ the Hungarian method 
\cite{kuhn1955hungarian} to match the clustering solution at each time to the 
previous result.

The identified water masses by static, AFFECT, and CESM schemes 
using AOLS with $L=3$ as the representation learning method are illustrated 
in Fig. \oldref{ocean} for  $t = 1,2,4,6$. The area colored with blue corresponds 
to the coast of south Africa and other islands in the target location. As we can 
see from the figure, all schemes are able to identify homogeneous water masses. 
However, the static subspace clustering and AFFECT schemes fail to properly 
detect the temporal changes in the formation of the green and orange water 
masses. In particular, the formation of the orange cluster evolves, as captured 
by the clustering results of the CESM framework. Since the  CESM framework 
accounts for the underlying temporal behavior in the 
representation learning process and are able to infer appropriate smoothing 
factors, they are able to accurately keep track of the orange and green clusters 
across different time steps. Note that similarly to the results on synthetic and 
real-time motion segmentation datasets, the smoothing parameter of AFFECT 
was approximately $0.5$. In addition, compared to AFFECT, CESM framework is 
capable of a faster adaptation to the changes in the formation 
of the orange water mass from $t = 1$ to $t =2$.

The temperature and salinity averages for the water masses clustered by the 
CESM framework are shown in Table \oldref{oceant} where the results in top, middle 
and bottom for each cluster correspond to $t = 2,4,6$, respectively. A combination 
of these values, the geographic location of the clusters, and prior studies in 
\cite{arzeno2016outcome,pickard2016descriptive} suggest that the 
purple, orange, and yellow clusters corresponds to Agulhas currents, AAIW, 
and the circumpolar deep water masses, respectively.
\section{Conclusion} \label{sec:concl}
In this paper, we studied the problem of evolutionary subspace clustering that is concerned 
with organizing a collection of data points that lie on a union of low-dimensional temporally
evolving subspaces. Unlike existing evolutionary clustering frameworks, the evolving data 
in evolutionary subspace clustering is assumed to be self-expressive, i.e., each data point 
can be represented by a linear combination of other data points in the set. By relying on the 
self-expressiveness property, we proposed a non-convex optimization framework that
enables learning parsimonious representations of data points at each time step while taking
into account representations from the preceding time step. The proposed framework utilizes 
a convex evolutionary self-expressive model (CESM) to establish a relation between current 
and previous representations. Finding parameters of CESM leads to a non-convex optimization 
problem; we developed schemes that rely on alternating minimization to solve it approximately 
and to adaptively tune a smoothing parameter that is reflective of the rate of evolution, i.e., 
indicates to what extent the representations in consecutive time steps are similar to each other. 
Our extensive studies on both synthetic and real-world datasets illustrate that the proposed CESM 
framework outperforms state-of-the-art static subspace clustering and evolutionary clustering 
schemes in majority of scenarios in terms of both accuracy and running time. Furthermore, the 
smoothing parameter learned by the proposed framework is interpretable and reflective of  the 
``memory" of data representation in consecutive time steps.

As part of the future work, it would be of interest to extend the CESM framework to other subspace 
clustering algorithms, including the schemes that rely on finding low rank representations of data 
points. It is also valuable to exploit the theoretical foundation of subspace clustering to analyze the 
performance of the proposed frameworks, e.g., in the setting of rotating random subspaces that we 
considered in this paper. Finally, it would be of interest to develop more complex models for the 
evolutionary subspace clustering problem, e.g., by using neural networks as the parametric function 
or a matrix of smoothing parameters in place of the proposed convex evolutionary self-expressive 
model.

\bibliographystyle{ieeetr}\small
\bibliography{refs.bib}
\end{document}